\documentclass[10pt,twocolumn,letterpaper]{article}

\usepackage{cvpr}
\usepackage{times}
\usepackage{epsfig}
\usepackage{graphicx}
\usepackage{amsmath}
\usepackage{amssymb}
\usepackage[pagebackref=true,breaklinks=true,letterpaper=true,colorlinks,bookmarks=false]{hyperref}
\usepackage{ctable}
\usepackage{multirow}
\usepackage{pifont}
\usepackage{comment}
\usepackage{float}
\usepackage{booktabs}
\usepackage[font=small]{subfig}
\usepackage{tabulary,overpic,xcolor}
\usepackage[font=small]{caption}

\cvprfinalcopy 

\def\cvprPaperID{1213} 

\ifcvprfinal\fi
\begin{document}

\title{Learning Video Representations from Correspondence Proposals}

\author{Xingyu Liu\thanks{Majority of the work done as an intern at Adobe Research.} \\ Stanford University
\and
Joon-Young Lee \\ Adobe Research
\and
Hailin Jin \\ Adobe Research
}

\maketitle

\begin{abstract}
Correspondences between frames encode rich information about dynamic content in videos. 
However, it is challenging to effectively capture and learn those due to their irregular structure and complex dynamics.
In this paper, we propose a novel neural network that learns video representations by aggregating information from potential correspondences.
This network, named CPNet, can learn evolving 2D fields with temporal consistency. In particular, it can effectively learn representations for videos by mixing appearance and long-range motion with an RGB-only input.
We provide extensive ablation experiments to validate our model. 
CPNet shows stronger performance than existing methods on Kinetics and achieves the state-of-the-art performance on Something-Something and Jester. We provide analysis towards the behavior of our model and show its robustness to errors in proposals.
\end{abstract}


\vspace{-2ex}
\section{Introduction}


Video modality can be viewed as a sequence of images evolving over time. A good model for learning video representations should be able to learn from both the static appearance of images as well as the dynamic change of images over time. 
The dynamic nature of video is described by temporal consistency, 
which says an object in one frame usually has its correspondence in other frames and its semantic features are carried along the way.
Analysis of these correspondences, either fine-grained or coarse-grained, can lead to valuable information for video recognition, such as how objects move or how viewpoints changes, which can further benefit high-level understanding tasks such as action recognition and action prediction.

Unlike static images where there is a standard representation learning approach of convolutional neural networks (CNNs),  the correspondence of objects in videos has entirely different pattern and is more challenging to learn. For example, the corresponding objects can be arbitrarily far away, may deform or change their pose, or may not even exist in other frames. Previous methods rely on operations within a local neighborhood (e.g. convolution) or global feature re-weighting (e.g. non-local means) for inter-frame relation reasoning thus cannot effectively capture correspondence: stacking local operations for wider coverage is inefficient or insufficient for long-range correspondences while global feature re-weighting fails to include positional information which is crucial for correspondence.

In this paper, we present a novel method of learning representations for videos from  correspondence proposals. Our intuition is that, the corresponding objects of a given object in other frames typically only occupy a limited set of regions, thus we need to focus on those regions during learning. In practice, for each position (a pixel or a feature), we only consider the few other positions that is most likely to be the correspondence.

\begin{figure}[t]
\centering
\includegraphics[width=0.47\textwidth]{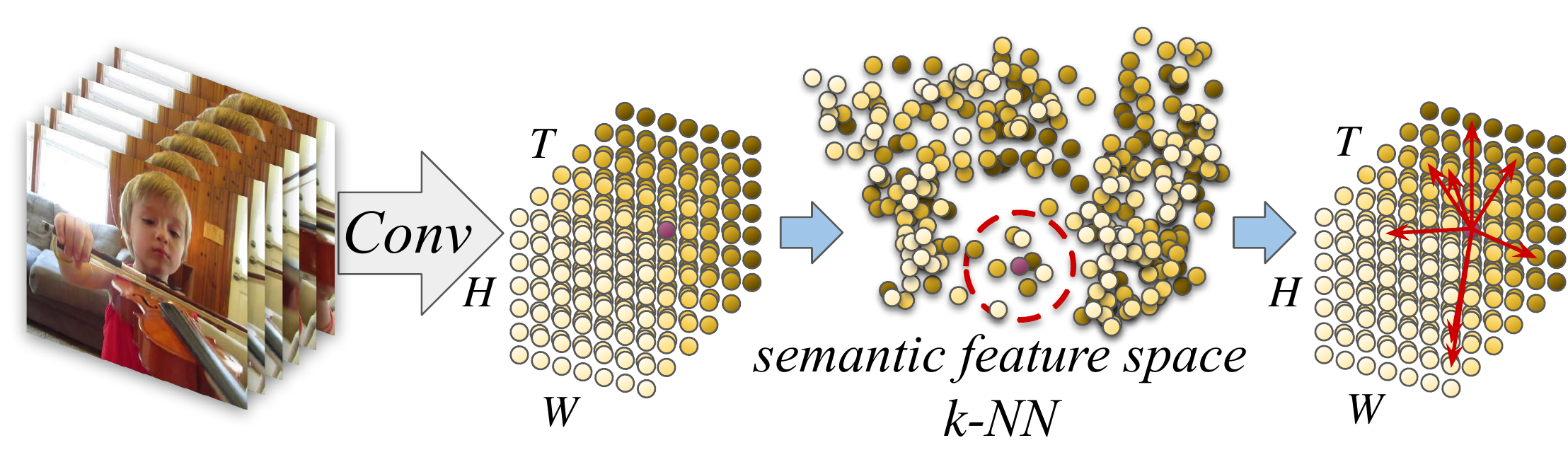}
\centering
\vspace{-2ex}
\caption{We view video representation tensor as a point cloud of features with $T\times H\times W$ points. For each point (e.g. the purple point), its $k$ potentially corresponding points are the $k$-NN in $C$-dimensional semantic space from other frames. Our CP module will learn and aggregate all these potential correspondences.}
\label{fig:teaser}
\vspace{-2ex}
\end{figure}

\begin{figure*}[t]
\centering
\includegraphics[width=1.0\textwidth]{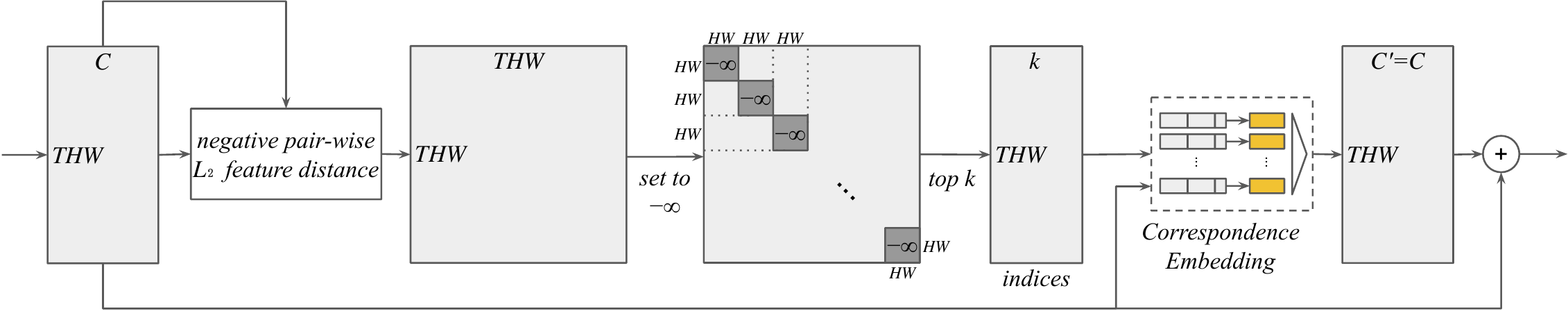}
\caption{CP module architecture. Gray boxes denote tensors, white boxes denote operators and orange boxes denote neural networks with trainable weights. The dashed box represents the Correspondence Embedding layer, whose  architecture is illustrated in detail in Figure \ref{fig:ce}.}
\label{fig:overall_arch}
\vspace{-2ex}
\end{figure*}

The key of our approach is a novel neural network module for video recognition named \emph{CP} module. This module views the video representation tensor as a point cloud in semantic space. As illustrated in Figure \ref{fig:teaser}, for every feature in video representation, the module finds and groups its $k$ nearest neighbors in other frames in the semantic space as potential correspondence. Each of the $k$ feature pairs is processed identically and independently by a neural network. Then max pooling is applied to select the strongest response.
  The module effectively learns a set of functions that embeds and selects the most interesting information among the $k$ pairs and encode the reason for such selection. The output of CP module is the encoded representation of correspondence, i.e. dynamic component in videos, and can be used in subsequent parts of an end-to-end architecture and other applications.

Ordered spatiotemporal location information is included in the CP module so that motion can be modelled. We integrate the proposed CP module into CNN so that both static appearance feature and dynamic motion feature of videos are mixed and learned jointly. We name the resulting deep neural network \emph{CPNet}. We constructed a toy dataset and showed that CPNet is the only RGB-only video recognition architecture that can effectively learn long-range motion. On real datasets, we show the robustness of the max pooling in CP module: it can filter out clearly wrong correspondence proposals and only select embeddings from reasonable proposals. 

We showcase CPNet in the application of video classification. We experimented with it on action recognition dataset Kinetics \cite{Kinetics} and compared it against existing methods. It beats previous methods and achieves leading performance. It also achieves state-of-the-art results among published methods on action-centric datasets Something-Something \cite{Something:Something} and Jester \cite{Jester} with fewer parameters. 
We expect that our CPNet and the ideas behind it can benefit video applications and research in related domains.






\section{Related Work}

\textbf{Representation Learning for Videos.} 
Existing approaches of video representation learning can generally be categorized by how dynamic component is modelled. The first family of approaches extract a global feature vector for each frame with a shared CNN and use recurrent neural nets to model temporal relation \cite{Long-Term:RCNN,Beyond:Short:Snippets}. Though recurrent architectures can efficiently capture temporal relations, 
it is harder to train and results in low performance on the latest benchmarks.
The second family of approaches learn dynamic changes from offline-estimated optical flow \cite{Two-stream:CNN,Two-stream:Fusion} or online learned optical flow \cite{TVLNet} with a separate branch of the network. The optical flow branch may share the same architecture as the static appearance branch. Though optical flow field bridges consecutive frames, the question of how to learn multiple evolving 2D fields is still not answered.

The third family of approaches use single-stream 3D CNN with RGB-only inputs and learn dynamic changes jointly and implicitly with static appearance \cite{C3D,I3D,Tran:ConvNet:search,Karpathy:Video:CNN,ARTNet,ECO}. These architectures are usually built with local operations such as convolution so cannot learn long-range dependencies. To address this problem, non-local neural nets (NL Nets) \cite{NLNet} was proposed. It adopted non-local operations where features are globally re-weighted by their pairwise feature distance. 
Our network consumes RGB-only inputs and explicitly computes correspondence proposals in a non-local fashion. Different from NL Net, our architecture focuses on only top correspondences and considers pairwise positional information, thus it can effectively learn not only appearance but also dynamic motion features.



\begin{figure*}[t]
\centering
\includegraphics[width=1.0\textwidth]{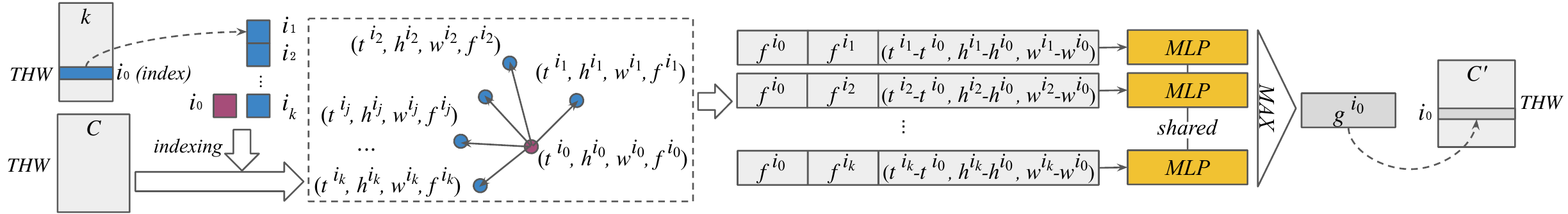}
\caption{Correspondence Embedding layer architecture.  $f^{i_j}$s are semantic vectors with length $C$ and the $i_j$-th row of input  $THW\times C$  feature tensor. $g^{i_0}$ is a semantic vector with length $C^\prime$ and the $i_0$-th row in the output  $THW\times C^\prime$  feature tensor. We made $C^\prime=C$ so that the output can be added back to the main stream CNN.  $t^{i_j},h^{i_j},w^{i_j}$ are the spatiotemporal normalized locations.}
\label{fig:ce}
\vspace{-2ex}
\end{figure*}

\textbf{Deep Learning on Unstructured Point Data. } The pioneering work of PointNet \cite{PointNet} proposed a class of deep learning methods on unordered point sets. The core idea is a symmetric function constructed with shared-weight deep neural networks followed by an element-wise max pooling. Due to the symmetry of pooling, it is invariant to the order of input points. This idea can also be applied to learning functions on generic orderless sets \cite{Deepset}.  
Follow-up work of PointNet++  \cite{PointNet++} extracts local features in local point sets within a neighborhood in the Euclidean space and hierarchically aggregates features. Dynamic graph CNN  \cite{Dynamic:Graph:CNN} proposed similar idea, the difference is that the neighborhood is determined in the semantic space and the neural network processes point pairs instead of individual points.
Inspired by these works, we treat correspondence candidates as an unordered set. Through a shared-weight MLP and max pooling, our network will learn informative representations about appearance and motion in videos.

\textbf{Deep Learning for Correspondence and Relation Reasoning.} Capturing relation is an essential task in computer vision and machine learning. A common approach to learn relation is letting extracted feature interact through a designed or learned function and discover similarity from the output. This is the general idea behind previous works on stereo matching \cite{Stereo:Matching:CNN:JMLR,Stereo:Matching:CNN:CVPR} and flow estimation \cite{FlowNet,FlowNet2,FlowNet3D}. The learned relation can also be used later in learning high-level semantic information such as video relational reasoning \cite{TRN} and visual question answering \cite{Relation:VQA}.
Compared to these works, we focus on learning video representation from long-range feature correspondences over time and space.

\section{Learning Correspondence Proposals}


Our proposed method is inspired by the following three properties of correspondences in videos:

1. \textbf{Corresponding positions have similar visual or semantic features.} 
This is the assumption underlying many computer vision tasks related to correspondence, such as image matching, relation reasoning or flow estimation.

2. \textbf{Corresponding positions can span arbitrarily long ranges, spatially or temporally.} In the case of fast motion or low frame rate, displacements along spatial dimensions can be large within small frame steps. Objects that disappear and then re-appear in videos across a long time can span arbitrarily long temporal range. 

3. \textbf{Potential correspondence positions in other frames are small in percentage.} 
Given a pixel/feature, usually only very small portion of pixels/features in other frames could be the potential correspondence. 
Other apparently dissimilar pixels/features can be safely ignored. 


A good video representation model should at least address the above three properties: it should be able to capture potential pixel/feature correspondence pairs at arbitrary locations and learn from the pairs. It poses huge challenges to the design of the deep architecture, since most deep learning methods work on regular structured data. Inspired by recent work on deep learning on point clouds \cite{PointNet,PointNet++,Dynamic:Graph:CNN} and their motion \cite{FlowNet3D}, we develop an architecture that addresses the above three properties. 

In this section, we first briefly review point cloud deep learning techniques and their theoretical foundations. Then we explain Correspondence Proposal (CP) module, the core of our architecture. Finally we describe how it is integrated into the entire deep neural network architecture. 

\subsection{Review of Point Cloud Deep Learning}

Qi et al. \cite{PointNet} recently proposed PointNet, a neural network architecture for deep learning in point clouds. 
Its theoretical foundation was proven in \cite{PointNet}: given 
a set of point clouds $\mathcal{X} \subseteq \{ \{x_1,x_2,\ldots, x_n\} \mid n \in \mathbb{Z}^+,  x_i \in [0,1]^d\}$
and any continuous set function  $f : \mathcal{X} \rightarrow \mathbb{R}^c$ w.r.t Hausdorff distance, symmetric function $g : \mathcal{X} \rightarrow \mathbb{R}^c$    
$$
g(x_1,x_2,\ldots,x_n) = \gamma \circ MAX \{\zeta(x_1), \zeta(x_2), \ldots, \zeta(x_n) \}
$$
can arbitrarily closely approximate $f$ on $\mathcal{X}$, where  $\zeta: \mathbb{R}^d\rightarrow\mathbb{R}^r$ and $\gamma: \mathbb{R}^r\rightarrow \mathbb{R}^c$ are two continuous functions and $MAX$ is the element-wise maximum operation. In practice, $\zeta$ and $\gamma$ are instantiated to be multi-layer perceptron (MLP) as learnable functions with universal approximation potential. The symmetry of max pooling ensures the output to be invariant to the ordering of the points.

While PointNet was originally proposed to learn geometric representation for 3D point clouds, it has been shown that the MLP can take mixed types of modalities as input to learn other tasks. For example, the MLP can take learned geometric representation and displacement in 3D Euclidean space as input to estimate scene flow \cite{FlowNet3D}.

\subsection{CP Module}

In this subsection, we explain the architectures of CP module. As illustrated in Figure \ref{fig:overall_arch}, the input and output to CP module are both video representation tensors with shape $THW\times C$, where $T$ denotes the number of frames, $H\times W$ denotes the spatial dimension and  $C$ denotes the number of channels. CP module treats the input video tensor as a point cloud of features with $THW$ points and accomplishes two tasks: 1) $k$-NN grouping; 2) correspondence embedding. 

\textbf{\boldmath{$k$}-NN grouping.} For each feature in the video representation tensor output by a CNN, CP module selects its $k$ most likely corresponding features in other frames. The selection is solely based on semantic similarity to ensure correspondence can be across arbitrarily long spatiotemporal ranges. Features within the same frame are excluded because temporal consistency should be between different frames.

The first step is to calculate the
semantic similarity of all features point pairs. 
We use the \textit{negative} $L_2$ distance as our similarity metric. 
It can be done efficiently with matrix multiply operations and produces a matrix of shape $THW\times THW$. 
The next step is to set the values of the elements in the $T$ diagonal block matrices of shape $HW\times HW$ to be $-\infty$. With this operation, the features within the same frame will be excluded from potential correspondences of each other. The final step is to apply an $\arg$ top-$k$ operation along the row dimension of the matrix. It outputs a tensor of shape $THW\times k$, where the $i$-th row are the indices of the $k$ nearest neighbors of the feature $i$. The workflow is shown in Figure \ref{fig:overall_arch}.

\textbf{Correspondence Embedding layer.} The goal of this layer is for each feature, learn a representation from its proposed correspondences.
The features' motion to their corresponding position in other frames can be learned during this process. 
The top-1 correspondence candidate can only give the information from one frame so it cannot capture the full correspondence information of the entire video. 
Besides, there may be more than one qualified correspondence candidates in a frame.
So we use a larger $k$, process $k$ pairs identically and independently, aggregate information from $k$ outputs. This is the general idea behind Correspondence Embedding layer, the core of our CP module.

Correspondence Embedding layer is located in the dashed box of Figure \ref{fig:overall_arch} and illustrated in detail in Figure \ref{fig:ce}. Suppose the spatiotemporal location and semantic vector of input feature $i_0$ is $(t^{i_0},h^{i_0},w^{i_0},f^{i_0})$, its $j$-th $k$-NN is $(t^{i_j},h^{i_j},w^{i_j},f^{i_j})$ where $j\in \{1,2,\ldots,k\}$. For each of the $k$ pairs, we pass the semantic vectors of two features, i.e. $f^{i_0}, f^{i_j} \in \mathbb{R}^C$, and their relative spatiotemporal displacements, i.e. $[t^{i_j}-t^{i_0},h^{i_j}-h^{i_0},w^{i_j}-w^{i_0}]\in\mathbb{R}^3$, to an MLP with shared weights. All three dimensions of the spatiotemporal locations, i.e. $t^{i_j}, h^{i_j}, w^{i_j}\in \mathbb{R}$, are normalized to $[0,1]$ from $[0,T)$, $[0,H)$ and $[0,W)$ before sent into MLP.  Then the $k$ outputs are aggregated by an element-wise max pooling operation to produce $g^{i_0} \in \mathbb{R}^C$, the semantic vector of output feature $i_0$. During the process, the most informative signals about correspondence, i.e. entangled representation from mixing displacement and two semantic vectors, can be selected from $k$ pairs and the output will implicitly encode motion information. Mathematically, the operation of Correspondence Embedding layer can be written as:
\begin{equation}\label{eq:ce:layer}
g^{i_0}=\underset{j\in \{1,2,\ldots,k\}}{MAX} \{ \zeta(f^{i_0}, f^{i_j}, t^{i_j}-t^{i_0},h^{i_j}-h^{i_0},w^{i_j}-w^{i_0}) \}
\end{equation}
where $\zeta$ is the function computed by MLP and $MAX$ is element-wise max pooling.

There are other design choices for Correspondence Embedding layer as well. For example, instead of sending both features directly to the MLP, one can first compute a certain distance between two features. 
However, as discussed in \cite{FlowNet3D}, sending both features into MLP is a more general form and yields better performance in motion learning.

\subsection{Overall Network Architecture}

Our CP module are inserted into CNN architecture and are interleaved with convolution layers, which enables the static image features extracted from convolution layers and correspondence signals extracted from CP module be mixed and learned jointly. 
Specifically, the CP modules are inserted into the ResNet \cite{ResNet} architectures and is located right after a residual block but before ReLU. 
We initialize the convolution part of our architecture with a pre-trained ImageNet model. The MLPs in CP modules are randomly initialized with MSRA initialization \cite{MSRA:init}, except for the gamma parameter of the last batch normalization layer \cite{batchnorm} being initialized with all zeros. This ensures identity mapping at the start of training so pre-trained model can be used.

In this paper, we only explore CP modules with $k$ nearest neighbors in other frames in $L_2$ \textit{semantic} space.
In general, however, the nearest neighbors of CP modules can be determined in other metric space as well, such as temporal-only space, spatiotemporal space or joint spatiotemporal-semantic space. We call such convolutional architecture inserted with generic CP module as CPNet.

\begin{table}[t!]
\small
\centering
\caption{Architectures for toy experiment}
\label{tab:toy}
\vspace{-2ex}
\setlength{\tabcolsep}{0.8pt}
\begin{tabular}{l|c|c|c|c}
layer & \begin{tabular}[c]{@{}c@{}}I3D NL \\ Net \cite{NLNet}\end{tabular} & \begin{tabular}[c]{@{}c@{}}ARTNet\\ \cite{ARTNet}\end{tabular} & \begin{tabular}[c]{@{}c@{}}TRN\\ \cite{TRN}\end{tabular} & \begin{tabular}[c]{@{}c@{}}C2D CPNet \\ (\textbf{ours}) \end{tabular} \\ \hline
conv$_1$ & $3\times3\times1$,16 & $3\times3\times3$,16 & $3\times3\times1$,16 & $3\times3\times1$,16 \\ \hline
 & NL block & - & - & CP module \\ \hline
conv$_2$ &  \begin{tabular}[c]{@{}c@{}}$1\times1\times3$,16\\ $3\times3\times1$,16\end{tabular}& \begin{tabular}[c]{@{}c@{}}SMART-\\ $3\times3\times3$,16\end{tabular} &  $3\times3\times1$,16 &  $3\times3\times1$,16 \\ \hline
& \begin{tabular}[c]{@{}c@{}}pooling, \\ fc \end{tabular} & \begin{tabular}[c]{@{}c@{}}pooling, \\ fc \end{tabular} &  \begin{tabular}[c]{@{}c@{}} pooling, temporal \\  relation, fc \end{tabular} & \begin{tabular}[c]{@{}c@{}}pooling, \\ fc \end{tabular} \\ 
\specialrule{.1em}{.05em}{.05em}
train & 27.8 & 26.8 & 27.1 & \textbf{97.9} \\ \hline
val & 26.4 & 25.9 & 26.9 & \textbf{97.4}
\end{tabular}
\end{table}

\begin{figure}[t]
\captionsetup{position=bottom}
\centering
\includegraphics[width=0.3\textwidth]{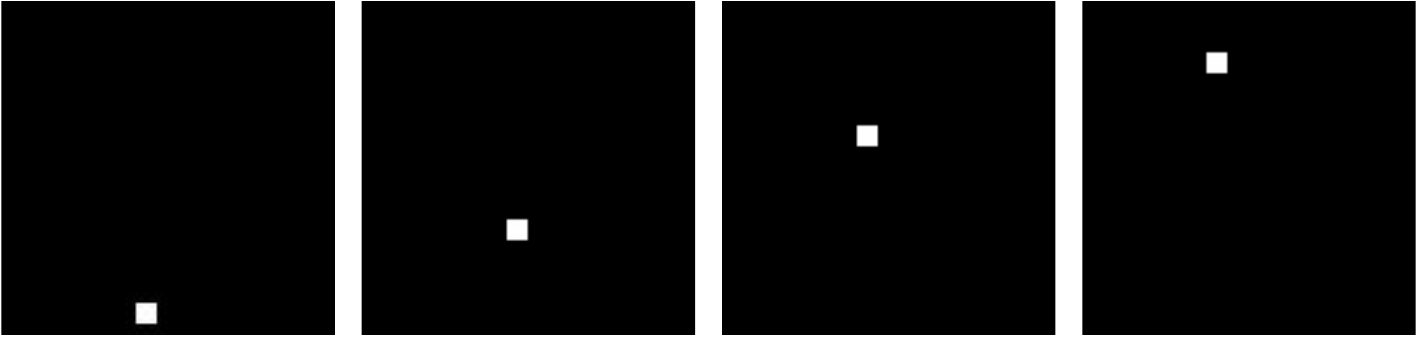}
\vspace{-2ex}
\caption{An ``up'' example in our toy dataset.} 
\label{fig:toy:viz}
\vspace{-1ex}
\end{figure}

\section{A Failing of Several Previous Methods}

We constructed a toy video dataset where previous RGB-only methods fail in learning long-range motion. Through this extremely simple dataset, we show the drawbacks of previous methods and the advantage of our architecture.

The dataset consists of videos of a $2\times 2$ white square moving on a black canvas. The videos have 4 frames and the spatial size is $32\times 32$. There are four labels of the moving direction of the square: ``left'', ``right'', ``up'' and ``down''. The square's moving distance per step is random between 7 and 9 pixels. The dataset has 1000 training and 200 validation videos, both have an equal number of videos for each label. Figure \ref{fig:toy:viz} illustrated an example of videos in our dataset.

We inserted the core module of several previous RGB-only deep architectures for video recognition, i.e. I3D NL Net \cite{NLNet}, ARTNet \cite{ARTNet}, TRN \cite{TRN}, as well as our CPNet, into a toy CNN with two $3\times3$ convolution layers. We listed the architectures used this experiment in Table \ref{tab:toy}. The convolution parts of all models have small spatial receptive fields of $5\times5$. The dataset-model settings are designed to simulate long-range motion situations where stacking convolution layers to increase receptive field is insufficient or inefficient.
No data augmentation is used. 

The training and validation results are listed in Table \ref{tab:toy}. Our model can overfit the toy dataset, while other models simply generate random guesses and fail in learning the motion.
It's easy to understand that ARTNet and TRN have insufficient convolution receptive fields to cover the step of the motion of the square. However, it's intriguing that NL Net, which should have a global receptive field, also fails. 

We provide an explanation as follows. Though the toy NL Net gets by the problem of insufficient convolution receptive fields, its NL block fails to include positional information thus can't learn long-range motion. 
However, it's not straightforward to directly add pairwise positional information to NL block without significantly increasing the memory and computation workload to an intractable amount.
Through this experiment, we show another advantage of our CPNet: by only focusing on top $k$ potential correspondences, memory and computation can be saved significantly thus allow positional information and semantic feature be learned together with more a complicated method such as a neural network.

\section{Experiment Results}

To validate the choice of our architecture for data in the wild, we first did a sequence of ablation studies on Kinetics dataset \cite{Kinetics}. 
Then we re-implemented several recently published and relevant architectures with the same dataset and experiment settings to produce results as good as we can and compare with our results. 
Next, we experiment with very large models and compare with the state-of-the-art methods on Kinetics validation set.
Finally, we did experiments on action-centric datasets Something-something v2 \cite{Something:Something} and Jester v1 \cite{Jester} and report our results on both validation and testing sets. Visualizations are also provided to help the understanding of our architecture.

\begin{table}[t]
\footnotesize
\setlength{\tabcolsep}{4.5pt}
\centering
\caption{Architectures used in Kinetics experiments in Table \ref{tab:kinetics:results}\protect\subref{tab:kinetics:comparison}.}
\vspace{-1.5ex}
\label{tab:architectures}
\begin{tabular}{l|c|c|c}
\multirow{2}{*}{layer} & \multirow{2}{*}{output size} & \multirow{2}{*}{\begin{tabular}[c]{@{}c@{}} C2D \\ baseline \end{tabular}} & \multirow{2}{*}{\begin{tabular}[c]{@{}c@{}} \textbf{CPNet (Ours)} \\ 6 CP modules\end{tabular}} \\
& & & \\ \hline
\multirow{2}{*}{conv$_1$} & \multirow{2}{*}{$56\times56\times8$} & \multirow{2}{*}{\begin{tabular}[c]{@{}c@{}} $7\times7$, 64, \\ stride 2, 2(, 1)\end{tabular}} & \multirow{2}{*}{\begin{tabular}[c]{@{}c@{}} $7\times7$, 64 \\ stride 2, 2\end{tabular}} \\ 
& & & \\ \hline
res$_2$  & $56\times56\times8$ & $\begin{bmatrix}
3\times 3, 64 \\
3\times 3, 64
\end{bmatrix} \times 2$ & $\begin{bmatrix}
3\times 3, 64 \\
3\times 3, 64
\end{bmatrix} \times 2$ \\ \hline
\multirow{3}{*}{res$_3$ } & \multirow{3}{*}{$28\times28\times8$} & \multirow{3}{*}{$\begin{bmatrix}
3\times 3, 128 \\
3\times 3, 128
\end{bmatrix} \times 2$} &  \multirow{3}{*}{\begin{tabular}[c]{@{}c@{}}$\begin{bmatrix}
3\times 3, 128 \\
3\times 3, 128 \\
\text{CP module}
\end{bmatrix} \times 2$\\ \end{tabular}} \\
&&&\\ 
&&&\\ \hline
\multirow{3}{*}{res$_4$ } & \multirow{3}{*}{$14\times14\times8$} & \multirow{3}{*}{$\begin{bmatrix}
3\times 3, 256 \\
3\times 3, 256
\end{bmatrix} \times 2$} & \multirow{3}{*}{\begin{tabular}[c]{@{}c@{}}$\begin{bmatrix}
3\times 3, 256 \\
3\times 3, 256 \\
\text{CP module}
\end{bmatrix} \times 2$\\ \end{tabular}} \\
&&&\\ 
&&&\\ \hline
\multirow{3}{*}{res$_5$ } & \multirow{3}{*}{$7\times7\times8$} & \multirow{3}{*}{$\begin{bmatrix}
3\times 3, 512 \\
3\times 3, 512
\end{bmatrix} \times 2$} & \multirow{3}{*}{\begin{tabular}[c]{@{}c@{}}$\begin{bmatrix}
3\times 3, 512 \\
3\times 3, 512 \\
\text{CP module}
\end{bmatrix} \times 2$\\ \end{tabular}} \\
&&&\\ 
&&&\\ \hline
& $1\times 1 \times 1$ & \multicolumn{2}{c}{global average pooling, fc $400$}  \\ 
\end{tabular}
\vspace{-3ex}
\end{table}

\subsection{Ablation Studies }
\label{sec:ablation:studies}
Kinetics \cite{Kinetics} is one of the largest well-labelled datasets for human action recognition from videos in the wild. Its classification task involves 400 action classes. It contains around 246,000 training videos and 20,000 validation videos.
We used C2D ResNet-18 as backbone for all ablation experiments. The architectures we used are derived from the last column of Table \ref{tab:architectures}. We included C2D baseline for comparison. 
We downsampled the video frames to be only 1/12 of the original frame rate and used only 8 frames for each clip. This ensures that the clip duration are long enough to cover a complete action while still maintain fast iteration of experiment. The single-clip single-center-crop validation results are shown in Table \ref{tab:kinetics:results}\subref{tab:kinetics:ablation:num}\subref{tab:kinetics:ablation:k}\subref{tab:kinetics:ablation:pos}.

\textbf{Ablation on the Number of CP modules.} We explored the effect of the number of CP modules on the accuracy. We experimented with adding one or two CP modules to the res$_4$ group, two CP modules to each of res$_3$ and res$_4$ groups, and two CP modules to each of  res$_3$, res$_4$ and $res_5$ groups. The results are shown in Table \ref{tab:kinetics:results}\subref{tab:kinetics:ablation:num}. As the number of CP modules increases, the accuracy gain is consistent.

\textbf{Ablation on \boldmath{$k$}.} We explored the the combination of training-testing time $k$ values and compared the results in Table \ref{tab:kinetics:results}\subref{tab:kinetics:ablation:k}. When $k$s are the same during training and testing, highest validation accuracy are achieved. It suggests that using different $k$ forces the architecture to learn different distribution and highest accuracy are achieved only when training and test distribution are similar. 

We also notice that the highest accuracy are achieved at a sweet point when both $k=8$. An explanation is that when $k$ is too small, CP module can't get enough correspondence candidates to select from; when $k$ is too large, clearly unrelated elements are also included and introduce noise.

\textbf{Ablation on the position of CP modules.} We explored effect of the position of CP modules. We added two CP modules to three different groups: res$_3$, res$_4$ and res$_5$, respectively. The results are shown in Table \ref{tab:kinetics:results}\subref{tab:kinetics:ablation:pos}. The highest accuracy are achieved when adding two CP modules to res$_4$ group. A possible explanation is that res$_3$ doesn't contain enough semantic information for finding correct $k$-NN while resolution of res$_5$  is too low ($7\times7$).

\begin{table*}[h]
\small
\caption{Kinetics datasets results for ablations and comparison with other prior works. The top-1/top-5 accuracies are shown.}
\label{tab:kinetics:results}
\vspace{-1.5ex}
\centering
\subfloat[number of CP modules]{\hspace{0ex}\begin{tabular}{l|cc}
model & top-1 & top-5\\ \hline
C2D & 56.9 & 79.5 \\ 
1 CP & 60.3  & 82.4  \\ 
2 CPs & 60.4 & 82.4 \\ 
4 CPs &  61.0 & \textbf{83.1}  \\ 
6 CPs & \textbf{61.1} & \textbf{83.1} \\ 
\end{tabular} \label{tab:kinetics:ablation:num}\hspace{2ex}} \hspace{1ex}
\subfloat[Ablation on CP module's $k$ values used in training and testing time.]{\begin{tabular}{l|l|c|c|c|c|c|c}
\multicolumn{2}{l|}{\multirow{2}{*}{\begin{tabular}[c]{@{}l@{}}top-1/top-5\\ accuracy\end{tabular}}} & \multicolumn{6}{c}{test} \\ \cline{3-8} 
\multicolumn{2}{l|}{}                                                                                & $k=1$ & $k=2$ & $k=4$  & $k=8$  & $k=16$ & $k=32$ \\ \hline
\multirow{6}{*}{train} & $k=1$ & \textbf{59.9/82.3} & 59.2/81.6 & 56.6/79.4 & 52.5/76.1 & 49.0/72.6 & 44.6/58.5 \\ \cline{2-8}
& $k=2$ & 59.1/81.8 & \textbf{60.2/82.5} & 59.6/81.8 & 56.9/80.1 & 53.0/77.1 & 48.9/73.5 \\ \cline{2-8}
& $k=4$ & 59.0/81.2 & 60.2/82.4 & \textbf{60.5/82.6} &   59.0/81.7   &   55.3/79.2   &   49.2/73.5   \\ \cline{2-8} 
& $k=8$ & 53.4/76.3 & 56.8/79.5 &  59.6/81.9   &   \textbf{60.7/82.8}   &  59.7/82.1  &   57.0/80.3   \\ \cline{2-8} 
& $k=16$ & 51.3/75.1 & 53.8/77.3 &  56.8/79.7   &   59.8/82.1   &    \textbf{60.6/82.8}  &  59.2/81.8    \\ \cline{2-8} 
& $k=32$ & 52.6/76.6 & 53.8/77.7 &  55.5/79.1    &   58.2/80.8   &   60.0/82.2   &   \textbf{60.4/82.4}  \\ 
\end{tabular} \label{tab:kinetics:ablation:k}} \\ \vspace{-2ex}
\subfloat[CP module positions]{\begin{tabular}{l|cc}
model & top-1 & top-5\\ \hline
C2D & 56.9 & 79.5 \\ 
res$_3$ & 60.4 & 82.4 \\ 
res$_4$ & \textbf{60.8} & \textbf{82.8} \\ 
res$_5$ & 59.2 & 81.6 \\
\end{tabular} \label{tab:kinetics:ablation:pos}}\hspace{3ex}
\subfloat[Kinetics validation accuracy of architectures in Table \ref{tab:architectures}. Clip length is 8 frames.]{
\begin{tabular}{l|p{1cm}p{1cm}|p{1cm}p{1cm}|p{1cm}p{1cm}|p{1cm}p{1cm}}
frame rate & \multicolumn{4}{c|}{1/12 of original frame rate} & \multicolumn{4}{c}{1/4 of original frame rate} \\ \hline
val configuration & \multicolumn{2}{c|}{1-clip, 1 crop} & \multicolumn{2}{c|}{25-clip, 10 crops} & \multicolumn{2}{c|}{1-clip, 1 crop} & \multicolumn{2}{c}{25-clip, 10 crops} \\ \hline
accuracy & \multicolumn{1}{c}{top-1} & \multicolumn{1}{c|}{top-5} & \multicolumn{1}{c}{top-1} & \multicolumn{1}{c|}{top-5} & \multicolumn{1}{c}{top-1} & \multicolumn{1}{c|}{top-5} & \multicolumn{1}{c}{top-1} & \multicolumn{1}{c}{top-5} \\ \specialrule{.1em}{.05em}{.05em}
C2D & \multicolumn{1}{c}{56.9} & \multicolumn{1}{c|}{79.5} & \multicolumn{1}{c}{61.3} & \multicolumn{1}{c|}{83.6} & \multicolumn{1}{c}{54.1} & \multicolumn{1}{c|}{77.4} & \multicolumn{1}{c}{60.8} & \multicolumn{1}{c}{83.3} \\ 
C3D \cite{C3D} & \multicolumn{1}{c}{58.3} & \multicolumn{1}{c|}{80.7} & \multicolumn{1}{c}{64.4} & \multicolumn{1}{c|}{85.8} & \multicolumn{1}{c}{55.0} & \multicolumn{1}{c|}{78.5} & \multicolumn{1}{c}{63.3} & \multicolumn{1}{c}{85.2} \\ 
NL C2D Net \cite{NLNet} & \multicolumn{1}{c}{58.6} & \multicolumn{1}{c|}{81.3} & \multicolumn{1}{c}{63.3} & \multicolumn{1}{c|}{85.1} & \multicolumn{1}{c}{55.3} & \multicolumn{1}{c|}{78.6} & \multicolumn{1}{c}{62.1} & \multicolumn{1}{c}{84.2} \\ 
ARTNet \cite{ARTNet} & \multicolumn{1}{c}{59.1} & \multicolumn{1}{c|}{81.1} & \multicolumn{1}{c}{65.1} & \multicolumn{1}{c|}{86.1} & \multicolumn{1}{c}{56.1} & \multicolumn{1}{c|}{78.7} & \multicolumn{1}{c}{64.2} & \multicolumn{1}{c}{85.6} \\ 
\textbf{CPNet (Ours)} & \multicolumn{1}{c}{\textbf{61.1}} & \multicolumn{1}{c|}{\textbf{83.1}} & \multicolumn{1}{c}{\textbf{66.3}} & \multicolumn{1}{c|}{\textbf{87.1}} & \multicolumn{1}{c}{\textbf{57.2}} & \multicolumn{1}{c|}{\textbf{80.8}} & \multicolumn{1}{c}{\textbf{64.9}} & \multicolumn{1}{c}{\textbf{86.5}} \\ 
\end{tabular} \label{tab:kinetics:comparison}
}\vspace{-4ex}
\end{table*}

\subsection{Comparison with Other Architectures}

We compare our architecture with C2D/C3D baselines, C2D NL Networks \cite{NLNet} and ARTNet \cite{ARTNet}, on Kinetics.
We did two sets of experiments, with frame rate downsampling ratio of 12 and 4 respectively. Both experiment sets used 8 frames per clip. The settings enable us to compare the performance under both low and high frame rates.  The architecture used in the experiments
are illustrated in Table \ref{tab:architectures}. 
We experimented with two inference methods: 25-clip 10-crop with averaged softmax score as in \cite{ARTNet} and single-clip single-center-crop. The results are shown in Table \ref{tab:kinetics:results}\subref{tab:kinetics:comparison}. 

Our architecture outperforms C2D/C3D baselines by a significant margin, which proves the efficacy of CP module. It also outperforms NL Net and ARTNet given fewer parameters, further showing the superiority of our CPNet.

\subsection{Large Models on Kinetics}
\label{sec:large:kinetics}

\begin{table}[t]
\small
\setlength{\tabcolsep}{5pt}
\centering
\caption{Large RGB-only models on Kinetics validation accuracy. Clip length for NL Net and our CPNet is 32 frames.}
\vspace{-1.5ex}
\label{tab:kinetics:sota}
\begin{tabular}{l|c|c|c}
model & params (M)  & top-1 & top-5\\ \hline
I3D Inception \cite{I3D} & 25.0 & 72.1 & 90.3 \\
Inception-ResNet-v2 \cite{Activity:Kinetics:challenge:2017} & 50.9 & 73.0 & 90.9 \\ 
NL C2D ResNet-101 \cite{NLNet} & 48.2 & 75.1 & 91.7 \\ \hline
\textbf{CPNet} C2D ResNet-101 \textbf{(ours)} & \textbf{42.1} & \textbf{75.3} & \textbf{92.4}
\end{tabular}
\vspace{-1ex}
\end{table}

We train a large model with C2D ResNet-101 as backbone. We applied three phases of training where we progressively increase the number of frames in a clip from 8 to 16 and then to 32. We freeze batch normalization layers starting the second phase. During inference, we use 10-clip in time dimension, 3-crop spatially fully-convolutional inference. The results are illustrated in Table \ref{tab:kinetics:sota}.

Compared with large models of several previous RGB-only architectures, our CPNet achieves higher accuracy with fewer parameters. We point out that Kinetics is an appearance-centric dataset where static appearance information dominates the classification. We will show later that our CPNet has larger advantage on other action-centric datasets where dynamic component more important.

\begin{table*}[h]
\small
\caption{TwentyBN datasets results. Our CPNet outperforms all published results, with fewer number of parameters.} 
\centering
\vspace{-1.5ex}
\label{tab:something:jester:results}
\subfloat[Something-Something v2 Results]{\begin{tabular}{l|c|c|c|c|c}
\multirow{2}{*}{model} &params & \multicolumn{2}{c|}{val}              & \multicolumn{2}{c}{test}             \\ \cline{3-6} 
& (M) & top-1 & top-5 & top-1 & top-5 \\ \hline
Goyal et al. \cite{Something:Something} & 22.2 & 51.33 & 80.46 & 50.76 & 80.77 \\ 
MultiScale TRN \cite{TRN} & 22.8 & 48.80 &77.64 & 50.85 & 79.33 \\ 
Two-stream TRN \cite{TRN} & 46.4  &  55.52 & 83.06 & 56.24 & 83.15 \\ \hline
C2D Res18 baseline & 10.7 & 35.24 & 64.49 & - & - \\ 
C2D Res34 baseline & 20.3 & 39.64 & 69.61 & - & - \\ 
\textbf{CPNet} Res18, 5 CP \textbf{(ours)} & \textbf{11.3} & 54.08 & 82.10 & 53.31 & 81.00 \\ 
\textbf{CPNet} Res34, 5 CP \textbf{(ours)} & \textbf{21.0} & \textbf{57.65} & \textbf{83.95} & \textbf{57.57} & \textbf{84.26} 
\end{tabular}\label{tab:something}} \hspace{0.5ex}
\subfloat[Jester v1 Results]{\begin{tabular}{l|c|c | c}
\multirow{2}{*}{model} & params  &\multirow{2}{*}{val} & \multirow{2}{*}{test} \\ 
 & (M) & & \\ \hline
BesNet \cite{BesNet} & 37.8 &  - & 94.23 \\
 MultiScale TRN \cite{TRN} & 22.8 & 95.31 & 94.78 \\ 
 TPRN \cite{TPRN} & 22.0 & 95.40 & 95.34 \\
 MFNet \cite{MFNet}  & 41.1 & 96.68  & 96.22 \\ 
 MFF \cite{MFF} & 43.4 & 96.33 & 96.28 \\ \hline
C2D Res34 baseline & 20.3 & 84.73 & - \\
\textbf{CPNet} Res34, 5 CP \textbf{(ours)} & \textbf{21.0} & \textbf{96.70} & \textbf{96.56}
\end{tabular}\label{tab:jester}}
\vspace{-2ex}
\end{table*}

\subsection{Results on Something-Something}
\label{sec:something:something}

Something-Something \cite{Something:Something} is a recently released dataset for recognizing human-object interaction from video. It has 220,847 videos in 174 categories. This challenging dataset is action-centric and especially suitable for evaluating recognition of motion components in videos. For example its categories are in the form of "Pushing something from left to right". Thus solely recognizing the object doesn't guarantee correct classification in this dataset. 

We trained two different CPNet models with ResNet-18 and -34 C2D as backbone respectively. We applied two phases of training where we increase the number of frames in a clip from 12 to 24. We freeze batch normalization layers in the second phase. The clip length are kept to be 2s \footnote{There are space for accuracy improvement when using 48 frames.}. During inference, we use 6-crop spatially fully-convolutional inference. We sample 16 clips evenly in temporal dimension from a full-length video and compute the averaged softmax scores over $6\times16$ clips. The results are listed in Table \ref{tab:something:jester:results}\subref{tab:something}.

Our CPNet model with ResNet-34 backbone achieves the state-of-the-art results on both validation and testing accuracy. Our model size is less than half but beat Two-stream TRN \cite{TRN} by more than 2\% in  validation accuracy and more than 1\% testing accuracy. Our CPNet model with ResNet-18 also achieves competing results. With fewer than half parameters, it beats MultiScale TRN \cite{TRN} by more than 5\% in validation and more than 2\% in testing accuracy.
Besides, we also showed the effect of CP modules by comparing against respective ResNet C2D baselines. Although parameter size increase due to CP module is tiny, the validation accuracy gain is significant ($>$14\%).

\subsection{Results on Jester}

Jester \cite{Jester} is a dataset for recognizing hand gestures from video. It has 148,092 videos in 27 categories. This dataset is also action-centric and especially suitable for evaluating recognizing motion components in video recognition models. One example of its categories is "Turning Hand Clockwise": solely recognizing the static gesture doesn't guarantee correct classification in this dataset. 
We used the same CPNet with ResNet-34 C2D backbone and the same training strategy as subsection \ref{sec:something:something}. During inference, we use 6-crop spatially fully-convolutional inference. We sample 8 clips evenly in temporal dimension from a full-length video and compute the averaged softmax scores over $6\times8$ clips. The results are listed in Table \ref{tab:something:jester:results}\subref{tab:jester}.

Our CPNet model outperforms all published results on both validation and testing accuracy, while having the smallest parameter size. The effect of CP modules is also shown by comparing against ResNet-34 C2D baselines. Again, although parameter size increase due to CP module is tiny, the validation accuracy gain is significant ($\approx$12\%).

\begin{figure*}[h]
\captionsetup{position=bottom}
\centering
\subfloat[A video clip with label ``playing basketball'' from Kinetics validation set.]{ 
\includegraphics[width=1.0\textwidth]
{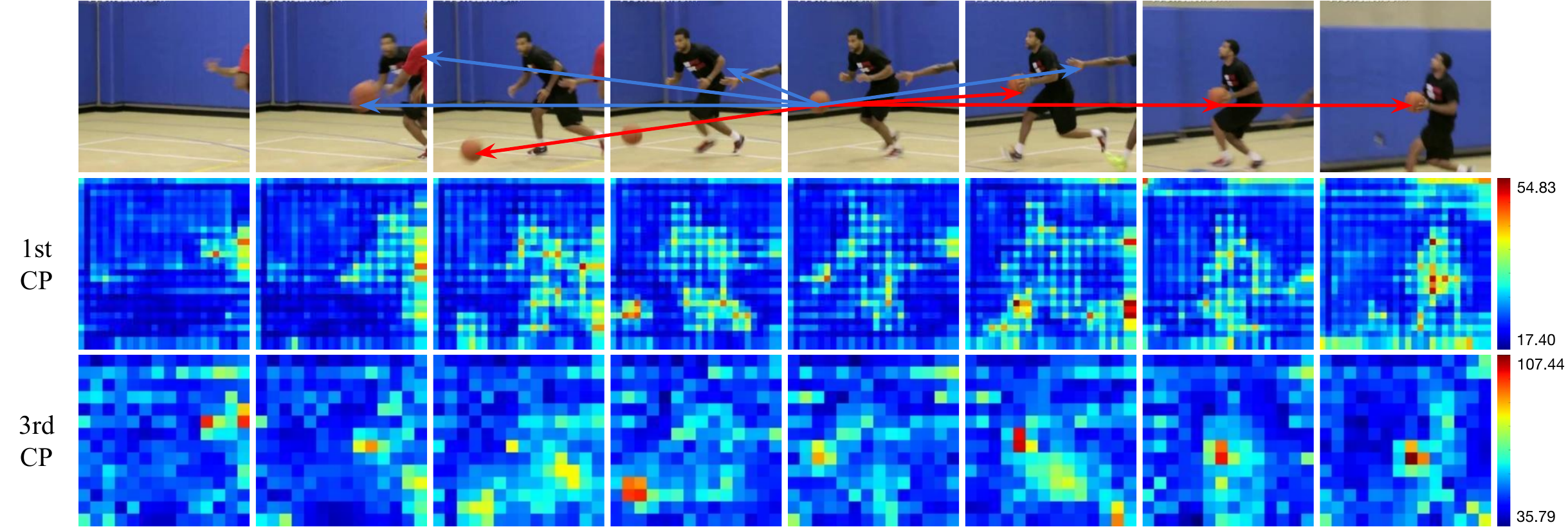} \label{fig:viz:kinetics}
} \vspace{-2ex} \\
\subfloat[A video clip with label ``Rolling something on a flat surface'' from Something-Something v2 validation set.]{
\includegraphics[width=1.0\textwidth]{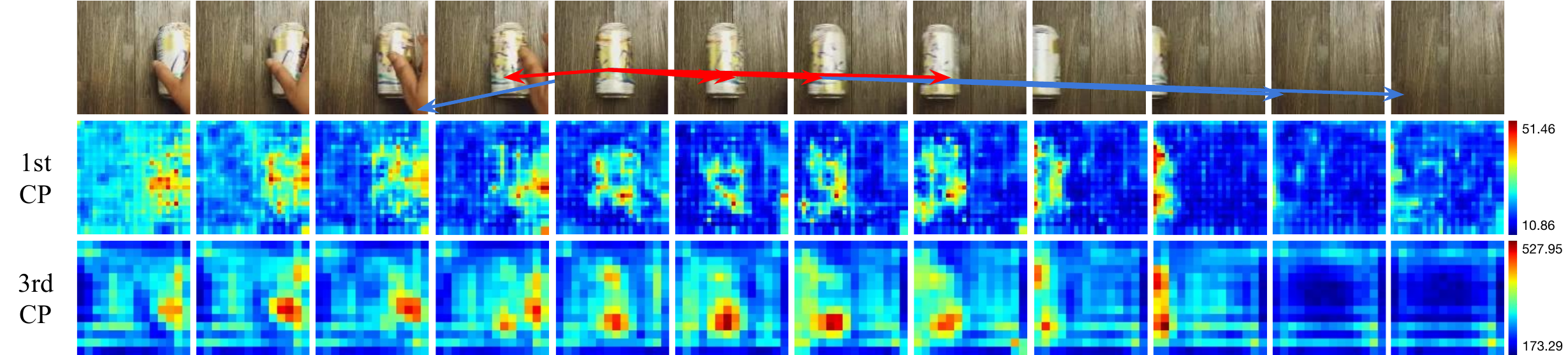} \label{fig:viz:something}
} \vspace{-2ex} \\
\subfloat[A video clip with label ``Thumb Up'' from Jester v1 validation set.]{
\includegraphics[width=1.0\textwidth]{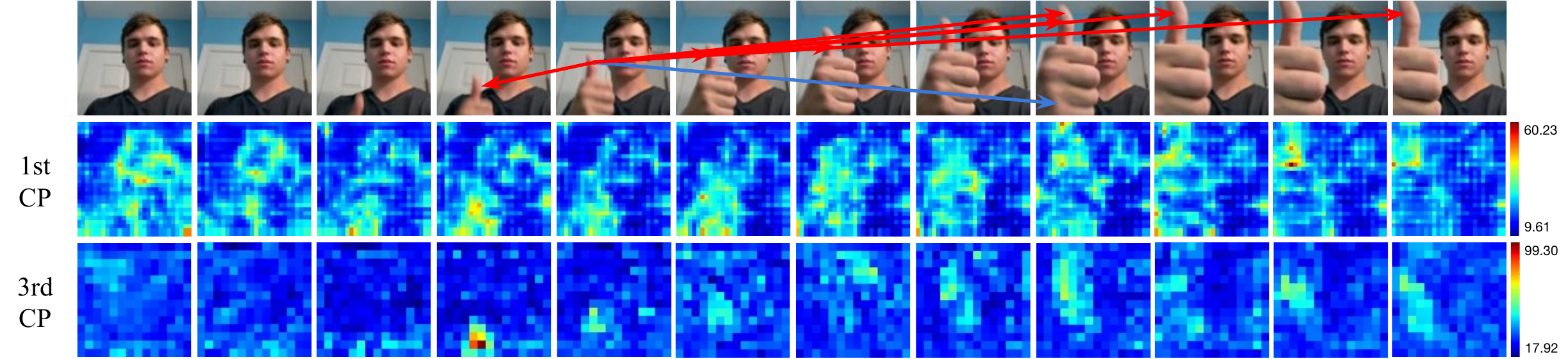} \label{fig:viz:jester}
} 
\caption{Visualization on our final models. The starting points of arrows are located at feature $i_0$. Arrows point to the $k$ proposed correspondences ($k=8$) of feature $i_0$. Proposed correspondences whose indices are in $\mathcal{A}^{i_0}$ (defined in Equation \eqref{eq:activate:neighbor}) are pointed by \textcolor{red}{red arrows} otherwise by \textcolor{blue}{blue arrows}. Feature changes after going through CP module are shown in heatmaps.} 
\label{fig:viz}
\end{figure*}

\subsection{Visualization}

To understand the behavior of CP module and demystify why it works,
we provide visualization in three aspects with the datasets used in previous experiments as follows.

\textbf{What correspondences are proposed?} We are interested to see whether CP module is able to learn to propose reasonable correspondences purely based on semantic feature similarity. As illustrated in Figure \ref{fig:viz}, in general CP module can find majority of reasonable correspondences. Due to $k$ being a fixed hyperparameter, its $k$-NN in semantic space may also include wrong correspondences.

\textbf{Which of proposed correspondences activate output neurons?} We are curious about CP module's robustness to wrong proposals. We trace which of the $k$ proposed correspondence pairs affect the value of output neurons after max pooling. Mathematically, let $g_c^{i_0}$ and $\zeta^{(i_0, i_j)}_c$ be the dimension $c$ of $g^{i_0}$ and $\zeta(f^{i_0}, f^{i_j}, t^{i_j}-t^{i_0},h^{i_j}-h^{i_0},w^{i_j}-w^{i_0})$ from Equation \eqref{eq:ce:layer} respectively, we are interested in the set
\begin{equation}\label{eq:activate:neighbor}
\mathcal{A}^{i_0} =\{j \in \{1,\ldots,k\} \mid \exists c\in \{1,\ldots,C\}, \zeta^{(i_0, i_j)}_c = g_c^{i_0} \}
\end{equation}
associated with a feature $i_0$, where $j$ not being in $\mathcal{A}^{i_0}$ means pair $(i_0,i_j)$ is entirely overwhelmed by other proposed correspondence pairs and thus filtered by max pooling when calculating output feature $i_0$. We illustrate $\mathcal{A}^{i_0}$ of several selected features in Figure \ref{fig:viz} and show that CP module is robust to incorrectly proposed correspondences. 

\textbf{How semantic feature map changes?} We show in Figure \ref{fig:viz} the heatmap of change in $L_1$ distance of the semantic feature map for each frame after going through CP module.
We found that CP modules make more changes to features that correspond to moving pixels. Besides, CP modules on a later stage focus more on the moving parts with specific semantic information that helps final classification.


\section{Discussion}

\subsection{Relation to Other Single-stream Architectures}

Note that since the MLPs in CP modules can potentially learn to approximate any continuous set functions, CPNet can be seen as a generalization of several previous RGB-only architectures for video recognition. 

CPNet can be reduced to a C3D  \cite{C3D} with kernel size $u\times v\times w$, if we set the $k$ of CP modules to be $uvw-1$, determine the $k$ nearest neighbors in spatiotemporal space with $L_1$ distance and let the MLP learn to compute inner product operation within the  $u\times v\times w$ neighborhood.

CPNet can also be reduced to an NL Net \cite{NLNet}, if we set the $k$ of CP modules to be maximum $THW-1$ and let the MLP learn to perform the same distance and normalization functions as the NL block.

CPNet can also be reduced to a TRN \cite{TRN}, if we put one final CP module at the end of C2D, determine the $k$ nearest neighbors in temporal-only space, and let the MLP learn to perform the same $g_\theta$ and $h_\phi$ functions defined in \cite{TRN}. 


\subsection{Pixel-level Motion vs. Feature-level Motion}

In two-stream architectures, motion in pixel level, i.e. optical flow fields, are first estimated before sent into deep networks. 
In contrast, CP modules captures motion in semantic feature level. We point out that, though CP module process positional information at a lower spatial resolution (e.g. $14\times14$), detailed motion feature can still be captured, since the semantic features already encode rich information within the receptive fields \cite{do:convnets:learn:correspondence}. 

In fact, migrating \textit{positional reasoning} from the original input data to semantic representation has contributed to several successes in computer vision research. 
For example, in the realm of object detection, moving the input and/or output of ROI proposal from the original image to the pooled representation tensor is the core of progress from RCNN \cite{RCNN} to Fast-RCNN \cite{Fast:RCNN} and to Faster-RCNN \cite{Faster:RCNN}; in the realm of flow estimation, successful architectures also calculate displacements within feature representations \cite{FlowNet,FlowNet2}.

\section{Conclusion}

In this paper, we presented a novel neural network architecture to learn representation for video. 
We propose a new CP model that computes $k$ correspondence proposals for each feature and feeds each of proposed pair to a shared neural network followed by max pooling to learn a new feature tensor. We show that the module can effectively capture motion correspondence information in videos. 
The proposed CP module can be integrated with most existing 
frame-based or clip-based 
video architectures. 
We show our proposed architecture achieves strong performance on standard video recognition benchmarks. In terms of future work, we plan to investigate this new architecture for problems beyond video classification.

\newpage

{\small
\bibliographystyle{ieee}
\bibliography{egbib}
}

 \section*{Supplementary}
\appendix

\section{Overview}

In this document, we provide more details to the main paper and show extra results on per-class accuracy and visualizations.

In section \ref{sec:kinetics:ablation}, we provide more details on the Kinetics/ResNet-18 ablation experiments (main paper section 5.1).
In section \ref{sec:kinetics:comp}, we provide more details on the baseline architectures in Kinetics/ResNet-18 comparison experiments (main paper section 5.2).
In section \ref{sec:kinetics:large}, we provide details on the CPNet architecture used  in Kinetics/ResNet-101 experiment (main paper section 5.3).
In section \ref{sec:something:jester}, we provide details on the architecture used  in Something-Something and Jester experiments (main paper section 5.4 and 5.5).
In section \ref{sec:per:class} we report the per-class accuracy of C2D model and our CPNet model on Something-Something and Jester datasets. Lastly in section \ref{sec:model:run:time} we provide time complexity of our model and in section \ref{sec:viz} we provide more visualization results on all three datasets.

\section{CPNet Architecture in  Kinetics/ResNet-18 Experiments}
\label{sec:kinetics:ablation}

Our CPNet is instantiated by adding a CP module after the last convolution layer of a residual group but before ReLU, as illustrated in Figure \ref{fig:instant_res18}. For Kinetics/ResNet-18 experiments in main paper section 5.1 and 5.2, each CP module has MLP with two hidden layers.
Suppose the number of channels of the input tensor of CP module is $C$. The number of channels of the hidden layers in the MLPs is then $[C/4, C/2]$. The number of nearest neighbors $k$ is set to 8 for the results in Table 3(a)(c)(d) of the main paper.  $k$ varies for the results in Table 3(b). The location of CP module is deduced from the last column of Table  \ref{tab:kinetics:comp:architectures} for different experiments in section 5.1 of the main paper.

\section{Baseline Architectures in Kinetics/ResNet-18 Comparison Experiment}
\label{sec:kinetics:comp}

\begin{figure}[t]
\centering
\includegraphics[width=0.47\textwidth]{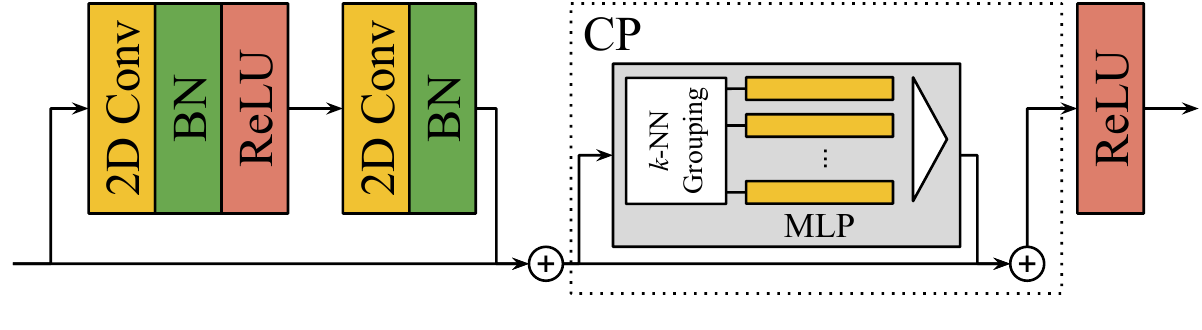}
\caption{CP module inserted into a residual group of ResNet-18 backbone.}
\vspace{-2ex}
\label{fig:instant_res18}
\vspace{-0ex}
\end{figure}

In Table \ref{tab:kinetics:comp:architectures}, we listed all the architectures used in Kinetics/ResNet-18 comparison experiments, as a supplementary to Table 2 of the main paper. C2D/C3D are vanilla 2D or 3D CNN. ARTNet is pulled directly from \cite{ARTNet}. It was designed to have the same number of parameters as its C3D counterpart. NL Net model is adapted from \cite{NLNet}, by adding an NL block at the end of each residual group of C2D ResNet-18. CPNet is instantiated in the same way as illustrated in Figure \ref{fig:instant_res18}. Combined with results in Table 3(d) of the main paper, our CPNet outperforms NL Net and ARTNet in terms of validation accuracy with fewer parameters, showing its superiority.

\begin{figure}[t]
\centering
\includegraphics[width=0.47\textwidth]{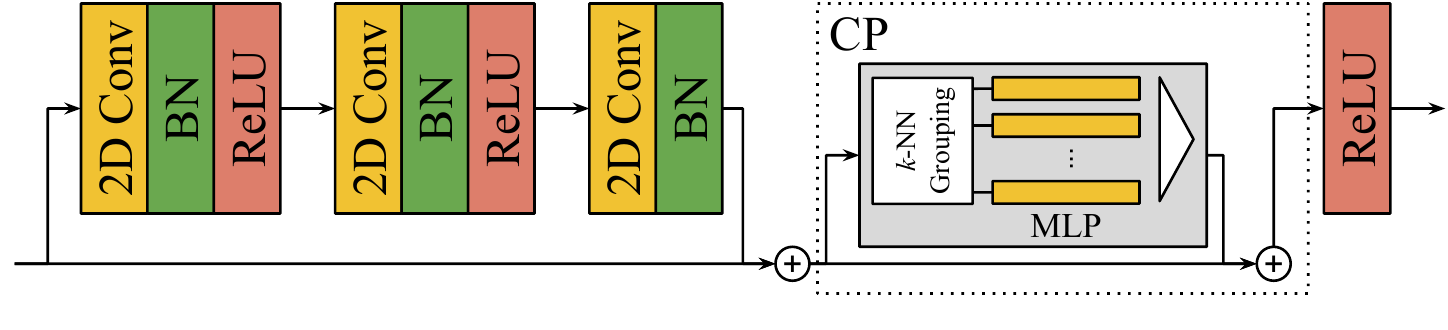}
\caption{CP module inserted into a residual group of ResNet-101 backbone.}
\vspace{-2ex}
\label{fig:instant_res101}
\vspace{-0ex}
\end{figure}

\begin{table*}[h]
\small
\centering
\caption{Complete Architectures used in Kinetics dataset comparison experiments.}
\vspace{-1.5ex}
\label{tab:kinetics:comp:architectures}
\begin{tabular}{l|c|c|c|c|c}
\multirow{2}{*}{layer} & \multirow{2}{*}{output size} & \multirow{2}{*}{C2D (C3D)} & \multirow{2}{*}{ARTNet \cite{ARTNet}} & \multirow{2}{*}{\begin{tabular}[c]{@{}c@{}}NL C2D Net \\ 6 NL blocks \cite{NLNet}\end{tabular}}  & \multirow{2}{*}{\begin{tabular}[c]{@{}c@{}} \textbf{CPNet (Ours)} \\ 6 CP modules\end{tabular}} \\
& & & & & \\ \hline
\multirow{2}{*}{conv1} & \multirow{2}{*}{$56\times56\times8$} & \multirow{2}{*}{\begin{tabular}[c]{@{}c@{}} $7\times7(\times3)$, 64, \\ stride 2, 2(, 1)\end{tabular}} & \multirow{2}{*}{\begin{tabular}[c]{@{}c@{}} SMART $7\times7\times3$, 64, \\ stride 2, 2, 1\end{tabular}} & \multirow{2}{*}{\begin{tabular}[c]{@{}c@{}} $7\times7$, 64, \\ stride 2, 2\end{tabular}} & \multirow{2}{*}{\begin{tabular}[c]{@{}c@{}} $7\times7$, 64, \\ stride 2, 2\end{tabular}} \\ 
& & & & & \\ \hline
res$_2$  & $56\times56\times8$ & $\begin{bmatrix}
3\times 3 (\times 3), 64 \\
3\times 3 (\times 3), 64
\end{bmatrix} \times 2$ & $\begin{bmatrix}
3\times 3 \times 3, 64 \\
\text{SMART }3\times 3 \times 3, 64
\end{bmatrix} \times 2$ & $\begin{bmatrix}
3\times 3, 64 \\
3\times 3, 64
\end{bmatrix} \times 2$ & $\begin{bmatrix}
3\times 3, 64 \\
3\times 3, 64
\end{bmatrix} \times 2$ \\ \hline
\multirow{3}{*}{res$_3$ } & \multirow{3}{*}{$28\times28\times8$} & \multirow{3}{*}{$\begin{bmatrix}
3\times 3 (\times 3), 128 \\
3\times 3 (\times 3), 128
\end{bmatrix} \times 2$} & \multirow{3}{*}{$\begin{bmatrix}
3\times 3 \times 3, 128 \\
\text{SMART }3\times 3 \times 3, 128
\end{bmatrix} \times 2$} & \multirow{3}{*}{\begin{tabular}[c]{@{}c@{}}$\begin{bmatrix}
3\times 3, 128 \\
3\times 3, 128 \\
\text{NL block}
\end{bmatrix} \times 2$\\ \end{tabular}} & \multirow{3}{*}{\begin{tabular}[c]{@{}c@{}}$\begin{bmatrix}
3\times 3, 128 \\
3\times 3, 128 \\
\text{CP module}
\end{bmatrix} \times 2$\\ \end{tabular}} \\
&&&&& \\ 
&&&&& \\ \hline
\multirow{3}{*}{res$_4$ } & \multirow{3}{*}{$14\times14\times8$} & \multirow{3}{*}{$\begin{bmatrix}
3\times 3 (\times 3), 256 \\
3\times 3 (\times 3), 256
\end{bmatrix} \times 2$} & \multirow{3}{*}{$\begin{bmatrix}
3\times 3 \times 3, 256 \\
\text{SMART }3\times 3 \times 3, 256
\end{bmatrix} \times 2$} & \multirow{3}{*}{\begin{tabular}[c]{@{}c@{}}$\begin{bmatrix}
3\times 3, 256 \\
3\times 3, 256 \\
\text{NL block}
\end{bmatrix} \times 2$\\ \end{tabular}} & \multirow{3}{*}{\begin{tabular}[c]{@{}c@{}}$\begin{bmatrix}
3\times 3, 256 \\
3\times 3, 256 \\
\text{CP module}
\end{bmatrix} \times 2$\\ \end{tabular}} \\
&&&&& \\ 
&&&&& \\ \hline
\multirow{3}{*}{res$_5$ } & \multirow{3}{*}{$7\times7\times8$} & \multirow{3}{*}{$\begin{bmatrix}
3\times 3 (\times 3), 512 \\
3\times 3 (\times 3), 512
\end{bmatrix} \times 2$} & \multirow{3}{*}{$\begin{bmatrix}
3\times 3 \times 3, 512 \\
3\times 3 \times 3, 512
\end{bmatrix} \times 2$} & \multirow{3}{*}{\begin{tabular}[c]{@{}c@{}}$\begin{bmatrix}
3\times 3, 512 \\
3\times 3, 512 \\
\text{NL block}
\end{bmatrix} \times 2$\\ \end{tabular}} & \multirow{3}{*}{\begin{tabular}[c]{@{}c@{}}$\begin{bmatrix}
3\times 3, 512 \\
3\times 3, 512 \\
\text{CP module}
\end{bmatrix} \times 2$\\ \end{tabular}} \\
&&&&&\\ 
&&&&&\\ \hline
& $1\times 1 \times 1$ & \multicolumn{4}{c}{global average pooling, fc $400$}  \\ \hline
\multicolumn{2}{c|}{params (M)} & 10.84 (31.81) & 31.81 & 10.88 & 10.86 \\
\end{tabular}
\vspace{-0.5ex}
\end{table*}

\begin{table}[t]
\small
\centering
\caption{CPNet Architectures used in Kinetics large model experiments.}
\vspace{-1.5ex}
\label{tab:kinetics:res101}
\begin{tabular}{l|c|c}
layer & output size & CPNet, 5 CP modules \\ \hline
\multirow{2}{*}{conv1} & \multirow{2}{*}{$56\times56\times8$} & \multirow{2}{*}{\begin{tabular}[c]{@{}c@{}} $7\times7$, 64, \\ stride 2, 2\end{tabular}} \\ 
& & \\ \hline
res$_2$  & $56\times56\times8$ & $\begin{bmatrix}
1\times 1, 64 \\
3\times 3, 64 \\
1\times 1, 256
\end{bmatrix} \times 3$ \\ \hline
\multirow{7}{*}{res$_3$ } & \multirow{7}{*}{$28\times28\times8$} & \multirow{7}{*}{\begin{tabular}[c]{@{}c@{}}$\begin{bmatrix}
1\times 1, 128 \\
3\times 3, 128 \\
1\times 1, 512
\end{bmatrix} \times 2$\\ $\begin{bmatrix}
1\times 1, 128 \\
3\times 3, 128 \\
1\times 1, 512 \\
\text{CP module}
\end{bmatrix} \times 2$  \end{tabular}} \\
&& \\ && \\ && \\ && \\ && \\ && \\ \hline
\multirow{7}{*}{res$_4$ } & \multirow{7}{*}{$14\times14\times8$} & \multirow{7}{*}{\begin{tabular}[c]{@{}c@{}}$\begin{bmatrix}
1\times 1, 256 \\
3\times 3, 256 \\
1\times 1, 1024 \\
\end{bmatrix} \times 20$ \\ $\begin{bmatrix}
1\times 1, 256 \\
3\times 3, 256 \\
1\times 1, 1024 \\
\text{CP module}
\end{bmatrix} \times 3$ \end{tabular}} \\
&& \\ && \\ && \\ && \\ && \\ && \\ \hline
\multirow{3}{*}{res$_5$ } & \multirow{3}{*}{$7\times7\times8$} & \multirow{3}{*}{\begin{tabular}[c]{@{}c@{}}$\begin{bmatrix}
1\times 1, 512 \\
3\times 3, 512 \\
1\times 1, 2048 \\
\end{bmatrix} \times 3$\\ \end{tabular}} \\
&&\\ 
&&\\ \hline
& $1\times 1 \times 1$ & \multicolumn{1}{c}{global average pooling, fc $400$} 
\end{tabular}
\vspace{-3ex}
\end{table}

\section{CPNet Architecture in Kinetics/ResNet-101 Experiment}
\label{sec:kinetics:large}

We listed CPNet architecture used in Kinetics/ResNet-101 experiment in Table \ref{tab:kinetics:res101}. Each residual group in ResNet-101 has three convolution layers. Our CPNet is instantiated by adding a CP module after the last convolution layer of a residual group but before ReLU, as  illustrated in Figure \ref{fig:instant_res101}. 
Suppose the number of channels of the input tensor of CP module is $C$. The number of channels of the hidden layers in the MLPs is then $[C/16, C/8]$. The number of nearest neighbors $k$ is set to 4.

We used five CP modules in the architecture. Two CP modules are in res$_3$ groups with spatial resolution of $28\times 28$ and the rest three are in res$_4$ groups with spatial resolution $14\times 14$. Such mixed usage of CP modules at residual groups of different spatial resolutions enables correspondence and motion in different semantic level to be learned jointly. We only listed the case of using 8 frames as input. For 32-frame input, all ``$8$'' in the second column of Table \ref{tab:kinetics:res101} should be replaced by $32$.

\section{Architecture used in Something-Something and Jester Experiments}
\label{sec:something:jester}

We listed the CPNet architectures used in Something-Something \cite{Something:Something} and Jester \cite{Jester} experiments in Table \ref{tab:something:jester}. CPNet is instantiated in the same way as illustrated in Figure \ref{fig:instant_res18}. 
Suppose the number of channels of the input tensor of CP module is $C$. The number of channels of the hidden layers in the MLPs is then $[C/4, C/2]$. The number of nearest neighbors $k$ is set to 12.

We used five CP modules in the architecture. Two CP modules are in res$_3$ groups with spatial resolution of $28\times 28$ and the rest three are in res$_4$ groups with spatial resolution $14\times 14$. We only listed the case of using 12 frames as input. For 24- or 48-frame input, all ``$12$'' in the second column of Table \ref{tab:kinetics:res101} should be replaced by $24$ or $48$.

\begin{table}[t]
\small
\centering
\setlength{\tabcolsep}{3.9pt}
\caption{CPNet Architectures used in Something-Something and Jester dataset experiments.}
\vspace{-1.5ex}
\label{tab:something:jester}
\begin{tabular}{l|c|c}
layer & output size & CPNet, 5 CP modules \\ \hline
\multirow{2}{*}{conv1} & \multirow{2}{*}{$56\times56\times12$} & \multirow{2}{*}{\begin{tabular}[c]{@{}c@{}} $7\times7$, 64, \\ stride 2, 2\end{tabular}} \\ 
& & \\ \hline
res$_2$  & $56\times56\times12$ & $\begin{bmatrix}
3\times 3, 64 \\
3\times 3, 64
\end{bmatrix} \times 3$ \\ \hline
\multirow{5}{*}{res$_3$ } & \multirow{5}{*}{$28\times28\times12$} & \multirow{5}{*}{\begin{tabular}[c]{@{}c@{}}$\begin{bmatrix}
3\times 3, 128 \\
3\times 3, 128
\end{bmatrix} \times 2$\\ $\begin{bmatrix}
3\times 3, 128 \\
3\times 3, 128 \\
\text{CP module}
\end{bmatrix} \times 2$  \end{tabular}} \\
&& \\ 
&& \\ 
&& \\ 
&& \\ \hline
\multirow{5}{*}{res$_4$ } & \multirow{5}{*}{$14\times14\times12$} & \multirow{5}{*}{\begin{tabular}[c]{@{}c@{}}$\begin{bmatrix}
3\times 3, 256 \\
3\times 3, 256 \\
\end{bmatrix} \times 3$ \\ $\begin{bmatrix}
3\times 3, 256 \\
3\times 3, 256 \\
\text{CP module}
\end{bmatrix} \times 3$ \end{tabular}} \\
&& \\ 
&& \\ 
&& \\ 
&& \\ \hline
\multirow{3}{*}{res$_5$ } & \multirow{3}{*}{$7\times7\times12$} & \multirow{3}{*}{\begin{tabular}[c]{@{}c@{}}$\begin{bmatrix}
3\times 3, 512 \\
3\times 3, 512
\end{bmatrix} \times 3$\\ \end{tabular}} \\
&&\\ 
&&\\ \hline
& $1\times 1 \times 1$ & \multicolumn{1}{c}{global average pooling, fc $174$ or fc $27$} 
\end{tabular}
\vspace{-2ex}
\end{table}

\section{Per-class accuracy of Something-Something and Jester models}
\label{sec:per:class}

\begin{figure}[h]
\centering
\includegraphics[width=0.47\textwidth]{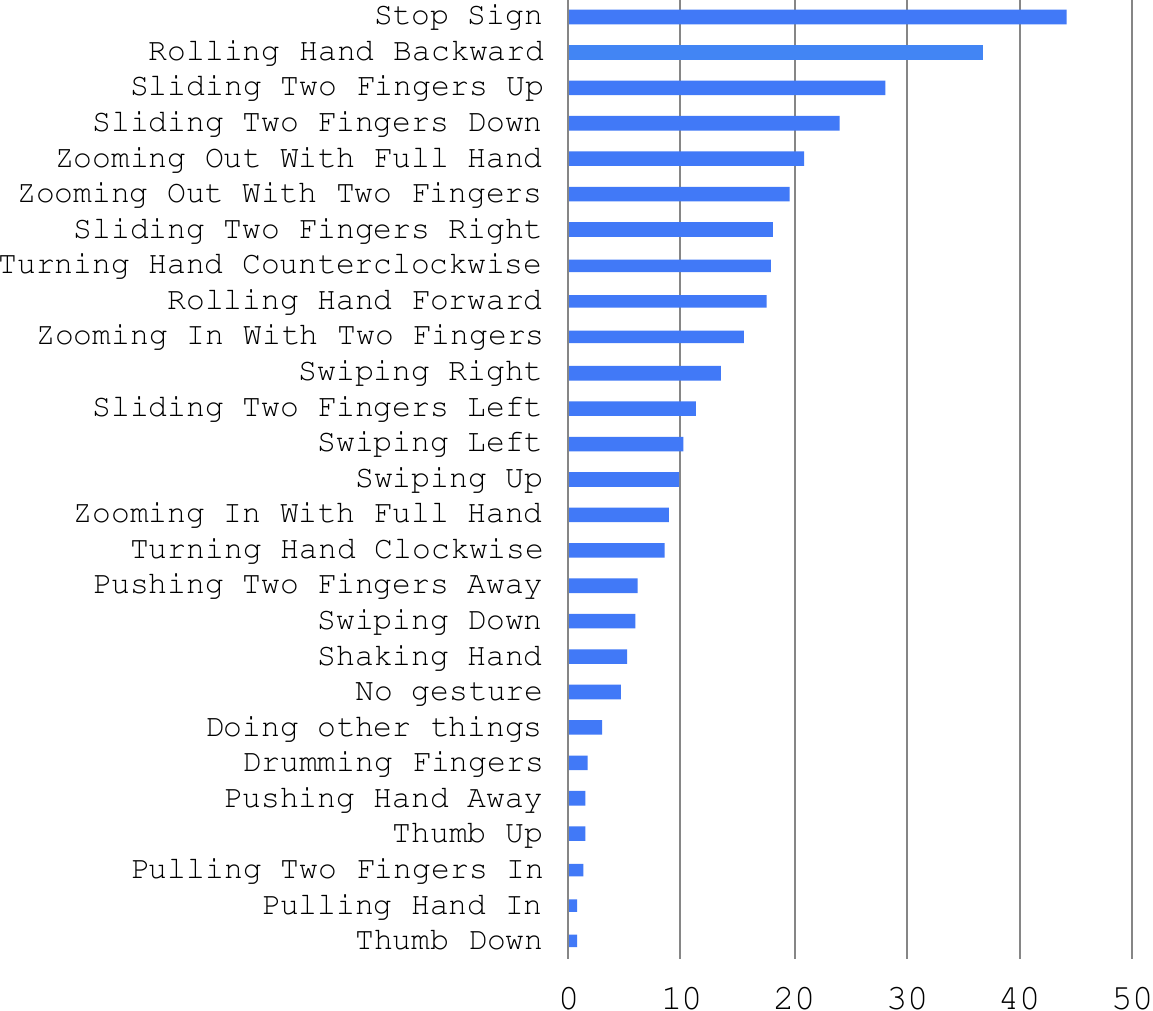}
\caption{Per-class top-1 accuracy gain in percentage on Jester v1 dataset due to CP module.}
\label{fig:per_class:jester}
\vspace{-2ex}
\end{figure}

\begin{figure*}[h]
\centering
\vspace{-5.5ex}
\includegraphics[width=0.615\textwidth]{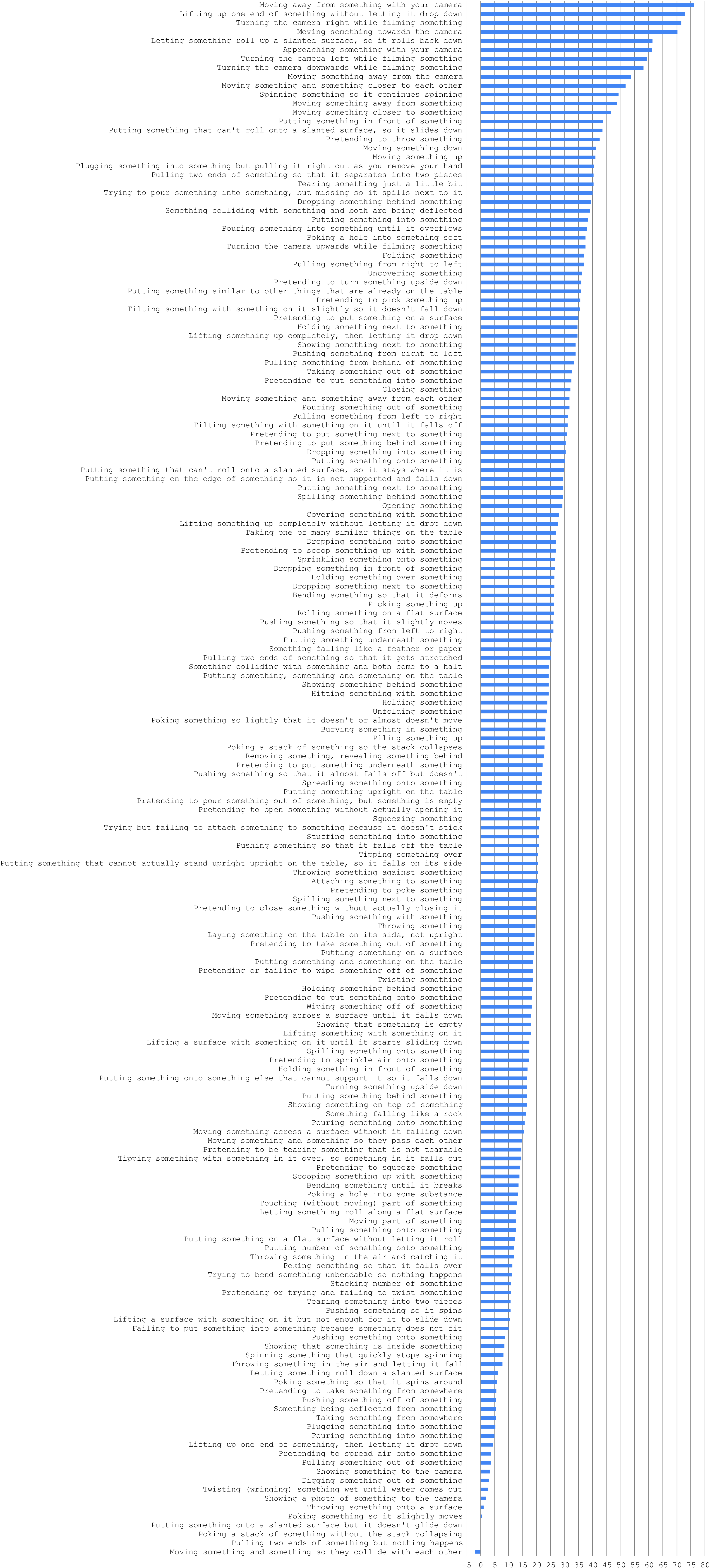}
\vspace{-2ex}
\caption{Per-class top-1 accuracy gain in percentage on Something-Something v2 dataset due to CP module.}\label{fig:per_class:something}
\end{figure*}

To understand the effect of CP module to the final performance, we provide the CPNet's per-class top-1 accuracy gain compared with the respective C2D baseline on Jester in Figure \ref{fig:per_class:jester} and Something-Something in Figure \ref{fig:per_class:something}.

We can see that categories that strongly rely on motion (especially in long-range) in videos typically have large accuracy improvement after adding CP module. On the other hand, categories that doesn't require reasoning motion to classify have little or negative gain in accuracy. The results coincide with our intuition that CP module effectively captures dynamic content of videos.

On Jester dataset \cite{Jester}, the largest accuracy improvements are achieved in categories that involve long-range spatial motion such as ``Sliding Two Fingers Up'', or long-range temporal relation such as ``Stop Sign''. At the same time, categories that don't even need multiple frames to classify, such as ``Thump Up'' or ``Thumb Down'', have the smallest accuracy gain.

On Something-Something dataset \cite{Something:Something}, the largest accuracy improvements are achieved in categories that involve long-range spatial motion such as ``Moving away from something with your camera'', or long-range temporal relation such as ``Lifting up one end of something without letting it drop down''. At the same time, categories that don't even need multiple frames to classify, such as ``Showing a photo of something to the camera'', have the smallest or negative accuracy gain .

\section{Model Run Time}
\label{sec:model:run:time}

In this section, we provide time complexity results of our model.
Our CP module can be very efficient in term of computation and memory, for both training and inference. 

During training, NL Net \cite{NLNet} computes a  $THW\times THW$ matrix followed by a row-wise softmax. The whole process is \textbf{differentiable} and all the intermediate values have to be stored for computing gradients during back propagation, which causes huge overhead in memory and computation. Unlike NL Net, our CP module's computation of a $THW\times THW$ matrix results in $k$ integers used for indexing, which is \textbf{non-differentiable}.  Thus CPNet doesn't compute gradients or store the intermediate values of the $THW\times THW$ matrix, a huge saving compared to NL Net and all  other works involving global attention.

\begin{figure}[t]
\centering
\includegraphics[width=0.47\textwidth]{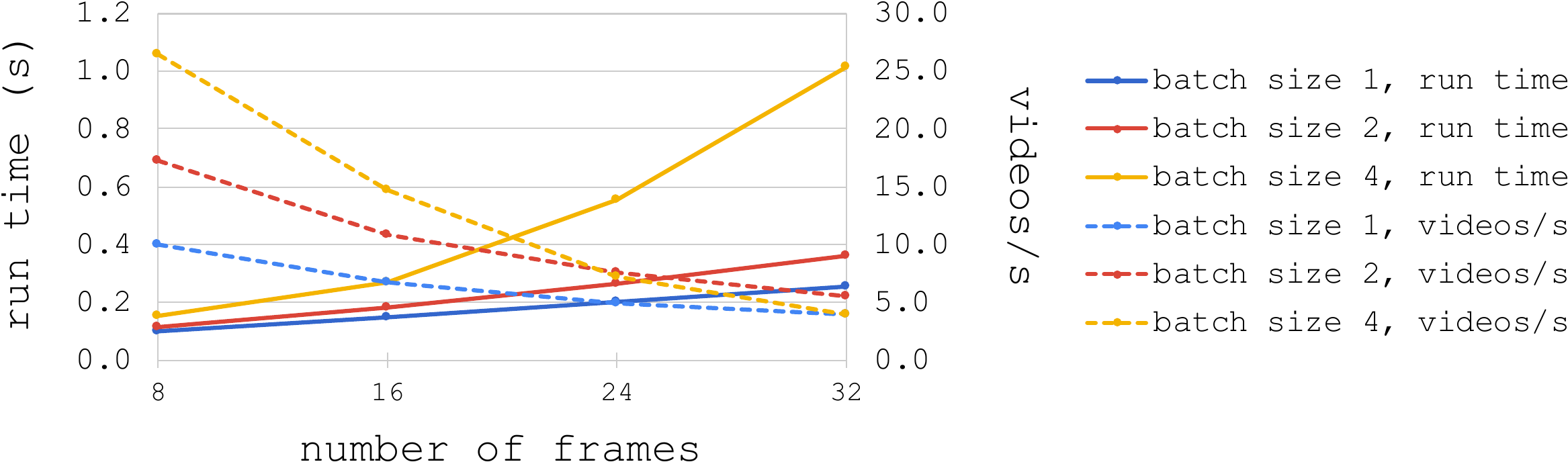}
\caption{Model run time (solid line) and number of video sequences per second (dashed line) of CPNet with ResNet-34 backbone and spatial size $112\times112$.}
\label{fig:run:time}
\vspace{-2ex}
\end{figure}

During inference, our CPNet is also efficient. We evaluate the inference time complexity of the CPNet model used in Jester v1 experiment. The spatial size is $112\times112$. The model backbone is ResNet-34. The computing platform is an NVIDIA GTX 1080 Ti GPU with Tensorflow and cuDNN. The model performances with various batch sizes and frame lengths are illustrated in Figure \ref{fig:run:time}. With batch size of 1, CPNet can reach processsing speed of 10.1 videos/s for frame length of 8 and 3.9 videos/s for frame length of 32. The number of videos that can be processed in a given time also increases as batch size increases.   

We point out that there exist other more efficient implementations of CP module.  In the main paper, we only presented the approach of the finding per-point $k$-NN in a point cloud via computing a pairwise feature distance matrix of size $THW\times THW$ followed by a row-wise $\arg$ top $k$, which has time complexity of $\mathcal{O} ((THW)^2\cdot (C+k))$. This is the most convenient way to implement in deep learning frameworks such as Tensorflow. 
However, when deployed on inference platforms, per-point $k$-NN  can be computed by much more efficient approaches with geometric data structures such as k-d tree \cite{kd-tree} or Bounding Volume Hierarchy (BVH) \cite{BVH} in $C$ dimensional space. The time complexity will then be $\mathcal{O} ((THW)\log(THW)\cdot (C+k))$, which includes both the construction and traversal of such tree data structures. Accelerating k-d tree or BVH on various platforms is an ongoing research problem in computer systems \& architectures community and is not the focus of our work. 

\section{More Visualizations}
\label{sec:viz}

\begin{figure*}[h]
\captionsetup{position=bottom}
\centering
\subfloat[A video clip with label ``ice skating'' from Kinetics validation set.]{
\includegraphics[width=1.0\textwidth]{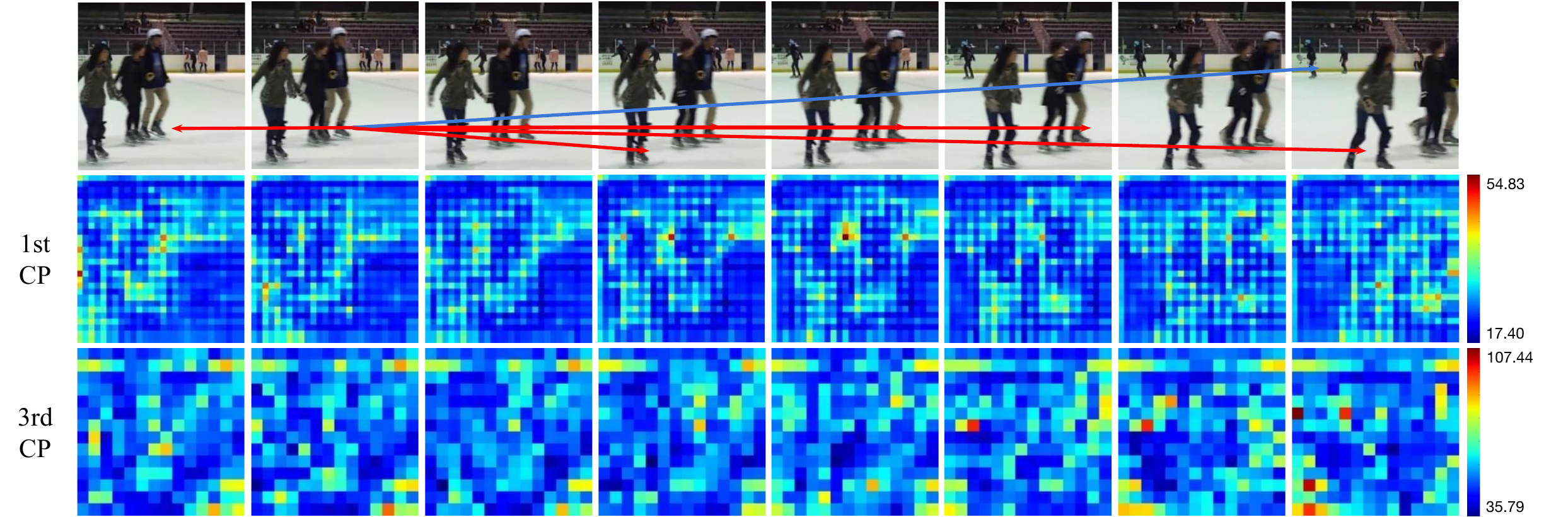} \label{fig:viz:kinetics:supp:ice_skating}
} 
\\
\subfloat[A video clip with label ``riding a bike'' from Kinetics validation set.]{
\includegraphics[width=1.0\textwidth]{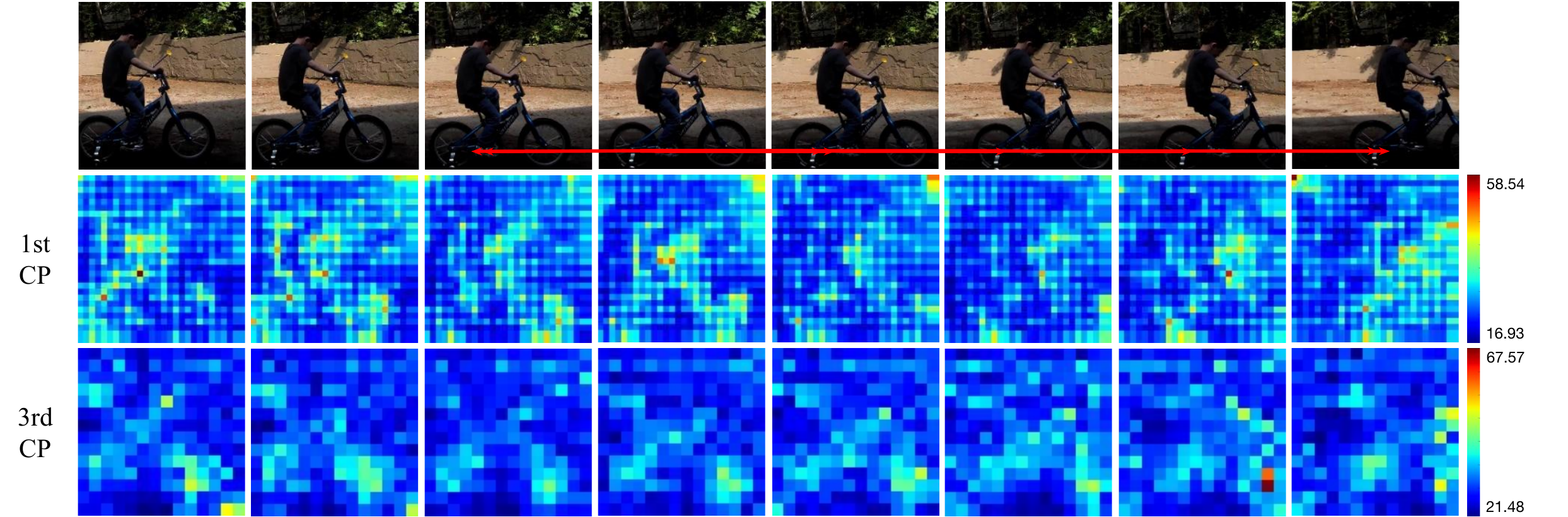} \label{fig:viz:kinetics:supp:riding_a_bike}
}  
\\
\subfloat[A video clip with label ``driving tractor'' from Kinetics validation set.]{
\includegraphics[width=1.0\textwidth]{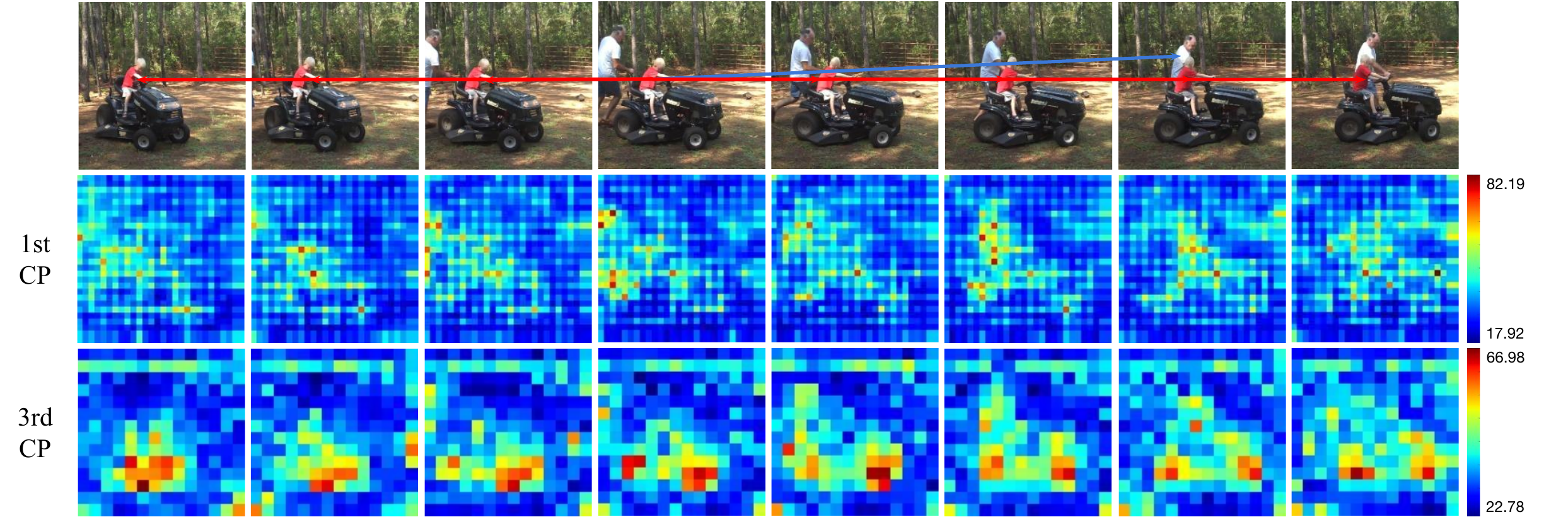} \label{fig:viz:kinetics:supp:driving_tractor}
} 
\caption{Additional Visualization on our final models on Kinetics dataset. Approach is the same as the main paper.} 
\label{fig:viz:kinetics:supp}
\end{figure*}

\begin{figure*}[h]
\captionsetup{position=bottom}
\centering
\subfloat[A video clip with label ``Turning something upside down'' from Something-Something v2 validation set.]{
\includegraphics[width=1.0\textwidth]{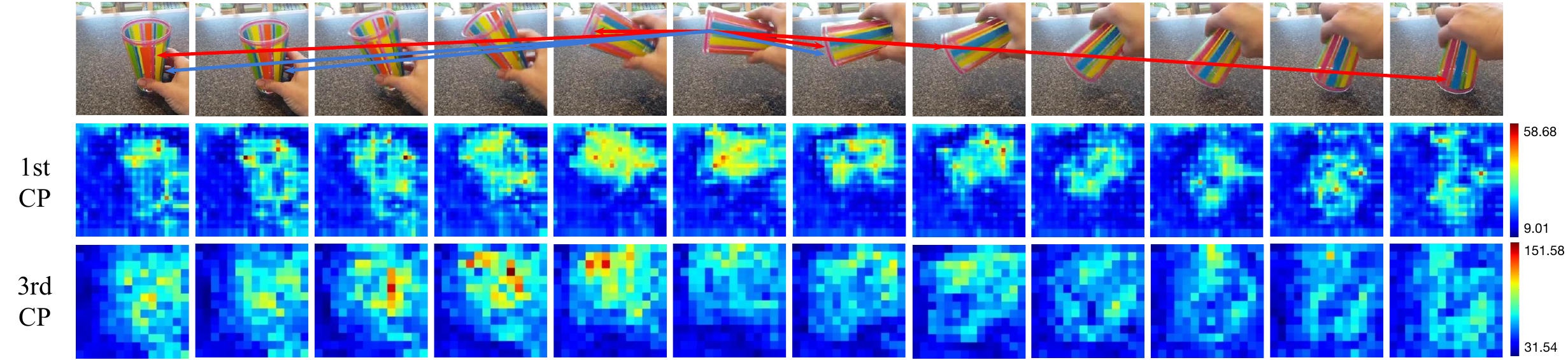} \label{fig:viz:something:supp:turning_something_upside_down}
} 
\vspace{-1ex} \\
\subfloat[A video clip with label ``Picking something up'' from Something-Something v2 validation set.]{
\includegraphics[width=1.0\textwidth]{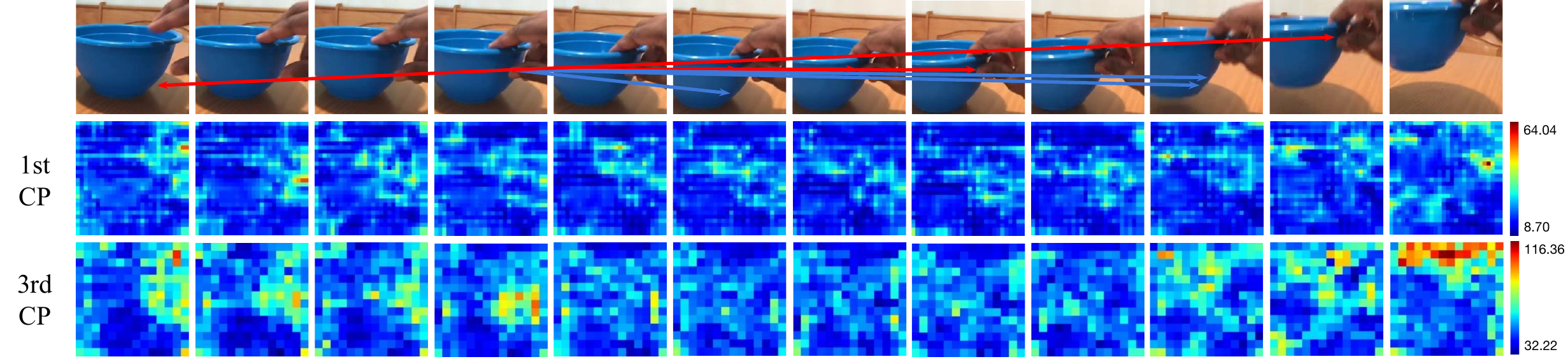} \label{fig:viz:something:supp:picking_something_up}
}  
\vspace{-1ex} \\
\subfloat[A video clip with label ``Moving something down'' from Something-Something v2 validation set.]{
\includegraphics[width=1.0\textwidth]{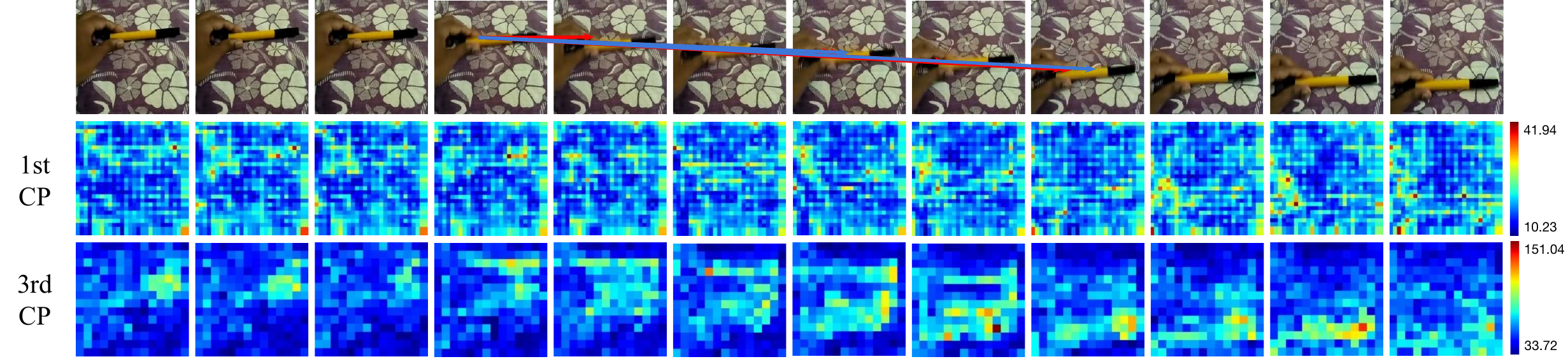} \label{fig:viz:something:supp:moving_something_down}
} 
\vspace{-1ex} \\
\subfloat[A video clip with label ``Dropping something next to something'' from Something-Something v2 validation set.]{
\includegraphics[width=1.0\textwidth]{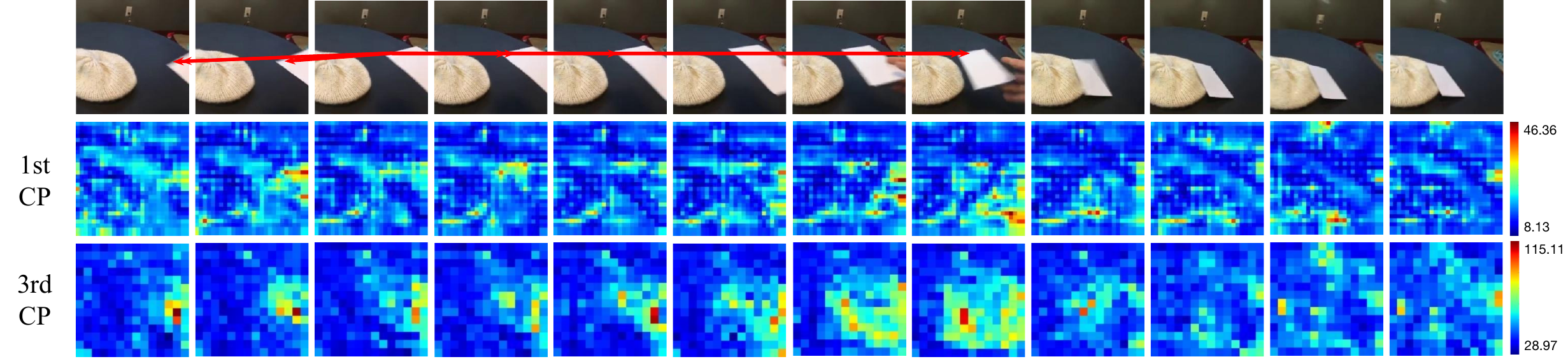} \label{fig:viz:something:supp:dropping_something_next_to_something}
} 
\caption{Additional Visualization on our final models on Something-Something v2 dataset. Approach is the same as the main paper.} 
\label{fig:viz:something:supp}
\end{figure*}

\begin{figure*}[h]
\captionsetup{position=bottom}
\centering
\subfloat[A video clip with label ``Drumming Fingers'' from Jester v1 validation set.]{
\includegraphics[width=1.0\textwidth]{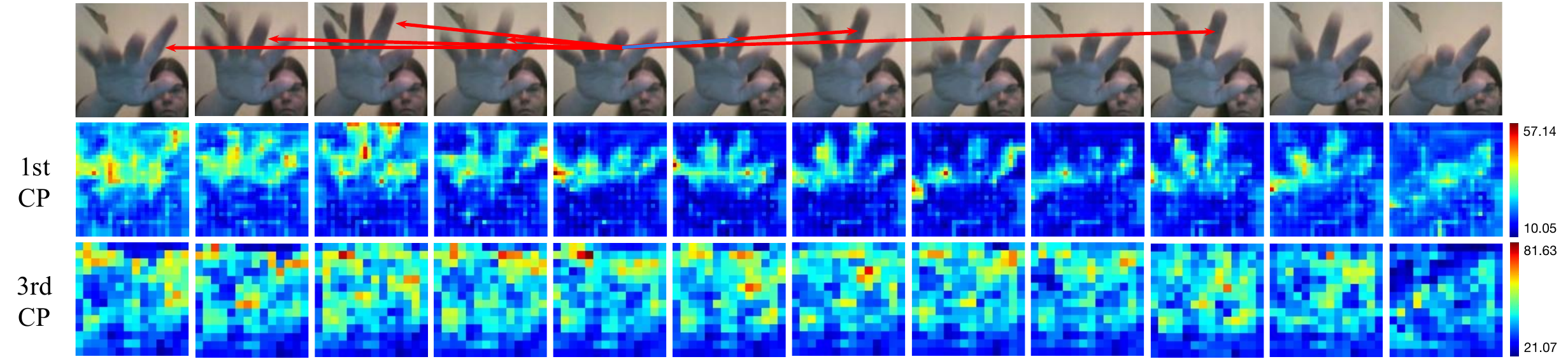} \label{fig:viz::suppjester:drumming_fingers}
} 
\vspace{-1ex} \\
\subfloat[A video clip with label ``Shaking Hand'' from Jester v1 validation set.]{
\includegraphics[width=1.0\textwidth]{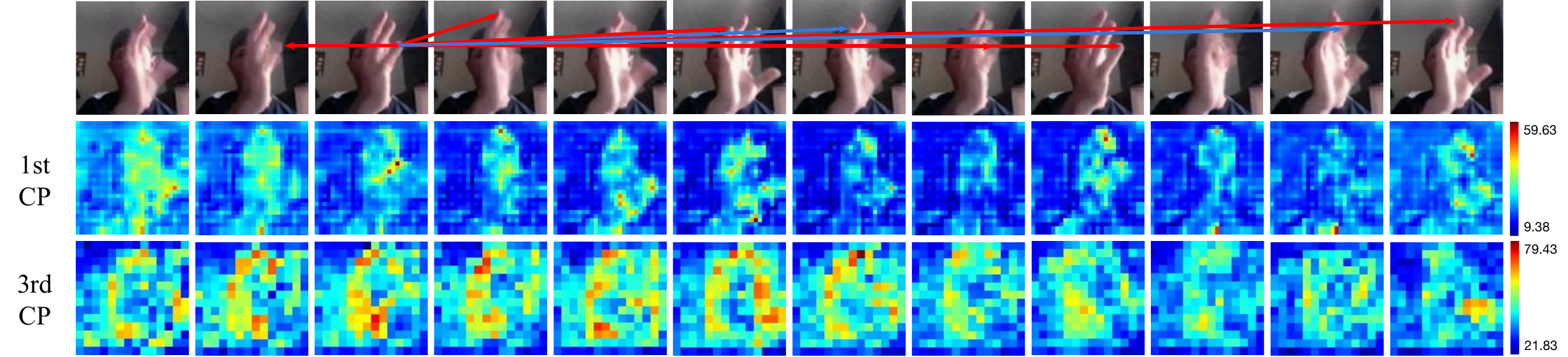} \label{fig:viz:jester:supp:shaking_hand}
}  
\vspace{-1ex} \\
\subfloat[A video clip with label ``Stop Sign'' from Jester v1 validation set.]{
\includegraphics[width=1.0\textwidth]{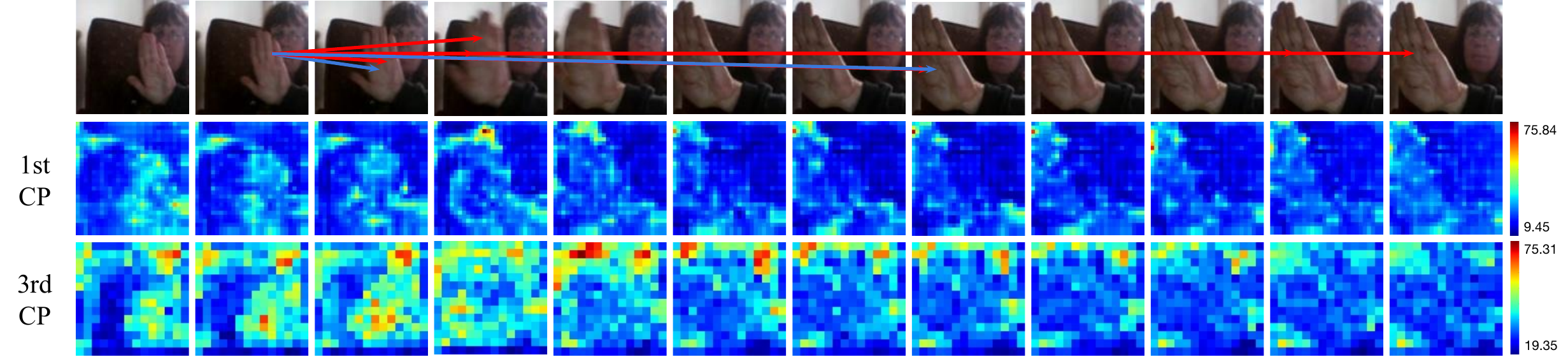} \label{fig:viz:jester:supp:stop_sign}
} 
\vspace{-1ex} \\
\subfloat[A video clip with label ``Pushing Two Fingers Away'' from Jester v1 validation set.]{
\includegraphics[width=1.0\textwidth]{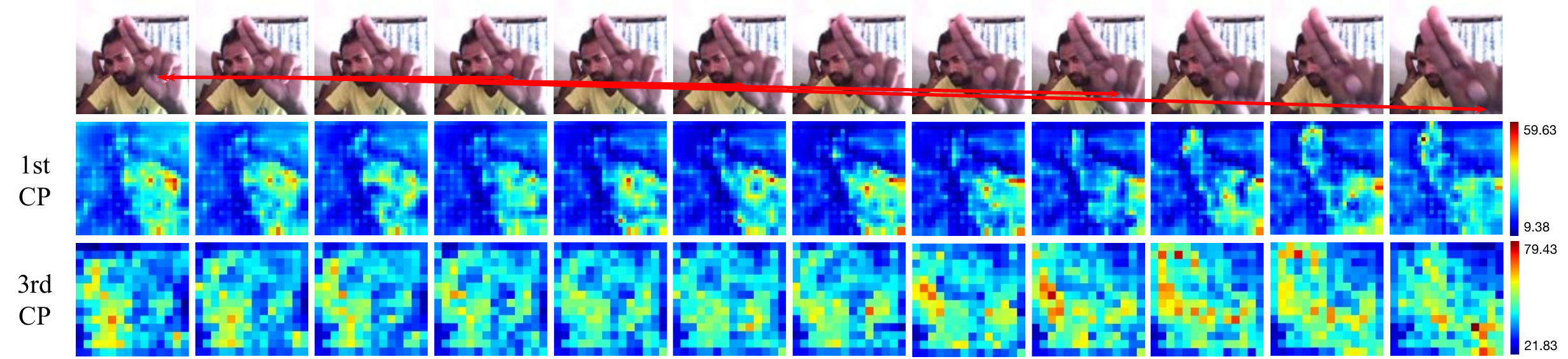} \label{fig:viz:jester:supp:pushing_two_fingers_away}
} 
\caption{Additional Visualization on our final models on Jester v1 dataset. Approach is the same as the main paper.} 
\label{fig:viz:jester:supp}
\end{figure*}

In this section, we provide more visualizations on examples from Kinetics \cite{Kinetics} in Figure \ref{fig:viz:kinetics:supp}, Something-Something \cite{Something:Something} in Figure \ref{fig:viz:something:supp} and Jester \cite{Jester} in Figure \ref{fig:viz:jester:supp}. They further show CP module's ability to propose reasonable correspondences and robustness to errors in correspondence proposal.

Despite what has been shown in the main paper, we also notice some negative examples. For example, in Figure \ref{fig:viz:kinetics:supp}\subref{fig:viz:kinetics:supp:ice_skating}, when proposing correspondences of the boy's left ice skate, CP module incorrectly proposed the a girl's left ice skate due to the two ice skates' visual features being too similar. CP module also didn't completely overwhelm this wrong proposal after max pooling. 
However, we notice that this wrong proposal is weak in the output signal: it only activates 3 out of 64 channels during max pooling which is acceptable.
We point out that such ``error'' could also be fixed in later stages of the network or even be beneficial for applications that require reasoning relations between similar but different objects.

\end{document}